\documentclass[letterpaper]{article} 
\usepackage{aaai2027}  
\usepackage[hyphens]{url}  
\usepackage{graphicx} 
\usepackage{natbib}  
\usepackage{caption} 
\usepackage{latexsym}
\usepackage[T1]{fontenc}
\usepackage[utf8]{inputenc}
\usepackage{microtype}
\usepackage{enumitem}
\usepackage{amssymb}
\usepackage{amsmath}
\usepackage{subcaption}
\usepackage{mdframed}
\usepackage[most]{tcolorbox}
\usepackage{tabularx}
\usepackage{multirow}
\usepackage{ragged2e}
\usepackage[table]{xcolor}
\usepackage{array}

\usepackage{algorithm}
\usepackage{algorithmic}

\usepackage{newfloat}
\usepackage{listings}
\DeclareCaptionStyle{ruled}{labelfont=normalfont,labelsep=colon,strut=off} 
\floatstyle{ruled}
\newfloat{listing}{tb}{lst}{}
\floatname{listing}{Listing}

\usepackage{booktabs}

\newtcolorbox{takeawaybox}[2][]{
    enhanced,
    attach boxed title to top left={xshift=3mm, yshift=-3mm},
    title=\textbf{#2},
    coltitle=white,
    colbacktitle=black,
    colframe=black,
    colback=blue!3!white,
    boxrule=0.7pt,
    arc=2mm,
    fonttitle=\small\sffamily,
    fontupper=\small,
    left=3mm, right=3mm, top=3mm, bottom=1mm,
    width=0.95\linewidth,
    center,
    #1
}

\title{Demystifying the Privacy-Utility Trade-off in LLM Interactions}
\author{
    Zhenhua Liu\textsuperscript{\rm 1}\equalcontrib,
    Zhanxu Xie\textsuperscript{\rm 2}\equalcontrib,
    Junjie Yu\textsuperscript{\rm 3,\rm 4}\equalcontrib,
    Tong Zhu\textsuperscript{\rm 5},
    Lijun Li\textsuperscript{\rm 5}\corresponding,
    Wenliang Chen\textsuperscript{\rm 1}\corresponding
}
\affiliations{
    \textsuperscript{\rm 1}Soochow University,
    \textsuperscript{\rm 2}Beihang University,
    \textsuperscript{\rm 3}Suzhou City University,\\
    \textsuperscript{\rm 4}Shanghai Key Lab of Intelligent Information Processing,
    \textsuperscript{\rm 5}Shanghai AI Lab,\\
    4065156@qq.com, wlchen@suda.edu.cn

}

\begin{document}

\maketitle

\begin{abstract}
The integration of Large Language Models into daily tasks relies on context-rich instructions, inevitably exposing sensitive user information. Current privacy-preserving methods typically employ context-agnostic static rules, causing severe utility degradation. However, the specific mechanisms governing how sanitization impacts downstream performance remain largely underexplored. To address this, we conduct a systematic analysis to deconstruct the privacy-utility trade-off, uncovering three underlying mechanisms: (1) \textbf{Context-Dependent Utility}, which first establishes when to sanitize by revealing that data value shifts from critical constraints to dispensable noise based on user intent; (2) \textbf{Strategic Adaptation}, which subsequently determines how to sanitize by dictating that the choice between removal and replacement depends on the task's reliance on factual integrity versus structural coherence; and (3) \textbf{Combinatorial Interplay}, which finally extends the protection scope by demonstrating that attributes form a semantic web of synergistic dependencies or antagonistic redundancies. Guided by these insights, we introduce an intent-driven local protection framework. By distilling a lightweight model \texttt{Veilmind-4B} to drive a dynamic extraction-sanitization-restoration pipeline, our approach reaches a low-leakage privacy point while preserving substantially higher response utility than existing privacy-oriented baselines, advancing the privacy-utility trade-off toward the Pareto frontier.
\end{abstract}


\section{Introduction}

Large Language Models (LLMs) have evolved into intelligent agents that seamlessly integrate into diverse user workflows \cite{brown2020language,xi2025rise}. To elicit high-quality responses, users must inevitably provide granular instructions rich in personal context and specific constraints \cite{salemi2024lamp, ouyang2022training}. Such detailed disclosure serves as the cornerstone for agents to accurately interpret intent and deliver personalized assistance \cite{wei2022chain}.

\begin{figure}[t]
    \centering
    \includegraphics[width=\linewidth]{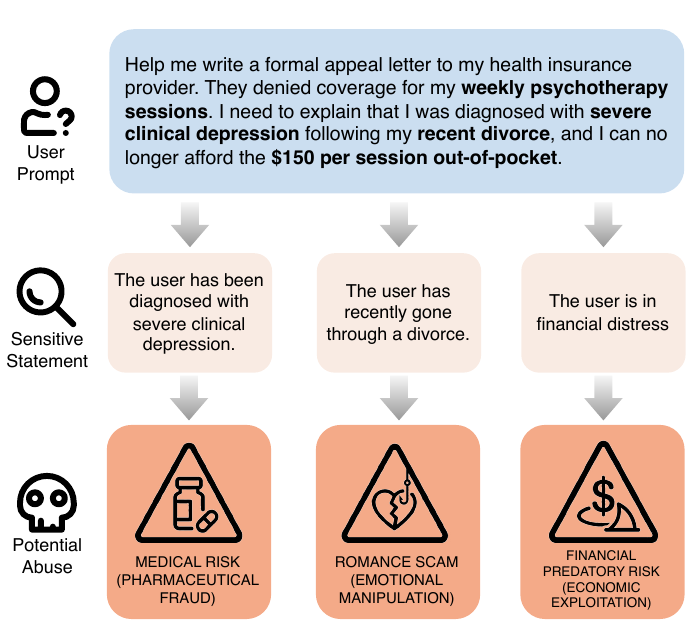}
    \caption{An illustrative example demonstrating how a user prompt can inadvertently disclose sensitive personal attributes, which may subsequently be exploited for targeted real-world abuse.}
    \label{fig:intro}
\end{figure}

However, high performance comes at a cost. To elicit precise assistance, users must disclose granular details that inevitably expose sensitive personal attributes. Crucially, data sensitivity is not static. It depends strictly on the user's intent and the specific task context. This aligns with the theory of contextual integrity \cite{nissenbaum2004privacy,nissenbaum2009privacy}. In practice, even seemingly innocuous prompts can enable the reconstruction of detailed user profiles \cite{staab2023beyond}. As depicted in Figure~\ref{fig:intro}, a routine task like drafting an insurance appeal necessitates the disclosure of clinical diagnoses and financial distress. Once transmitted to untrusted servers, this data risks being mined for detailed profiling and subsequently exploited for targeted real-world abuse \cite{carlini2021extracting,neel2023privacy,wang2023decodingtrust}.

To mitigate these risks, prior research has primarily relied on static sanitization rules or generic heuristics \cite{lison2021anonymisation,edemacu2025privacy}. While effective at reducing privacy risks, they frequently compromise the utility of the response \cite{feyisetan2020privacy,chowdhury2025prepsilonepsilonmptsanitizingsensitiveprompts}. The critical limitation lies in their indiscriminate treatment of sensitive attributes. Such methods fail to evaluate the marginal contribution of specific details relative to the user's goal, often stripping away essential context alongside non-essential noise \cite{mireshghallah}. Consequently, the field lacks a granular understanding of how varying degrees of sanitization impact model performance, leaving the underlying mechanisms of \textbf{the privacy-utility trade-off} largely unexplored.

To address this gap, we conduct a systematic empirical analysis to deconstruct the privacy-utility trade-off. Unlike prior black-box approaches, we move beyond static heuristics to map the underlying decision boundaries, answering three fundamental questions:

\begin{itemize}[leftmargin=*]
    \item \textbf{When to Sanitize?} We reveal that the functional value of sensitive information is contingent on the task context. We observe distinct dynamics ranging from functional coupling where attributes act as critical constraints to informational redundancy where they serve as dispensable noise. Consequently, decisions on necessity must be grounded in this contextual utility.

    \item \textbf{How to Sanitize?} We demonstrate that optimal protection hinges on prioritizing either factual integrity or structural coherence. Objective tasks demand removal to prevent false premises while interactive scenarios require replacement to serve as conversational anchors. Thus, the choice of strategy is strictly dictated by the task's reliance on factual versus structural validity.

    \item \textbf{What Scope to Sanitize?} We demonstrate that sensitive attributes do not function in isolation but form a complex semantic web. This structure creates either synergistic bundles essential for coherence or antagonistic redundancy capable of leaking information. Consequently, protection strategies must transcend individual evaluation to address the scope of these combinatorial dependencies.
\end{itemize}

Guided by these insights, we propose an intent-driven local protection framework. To endow a lightweight local model with the advanced reasoning capabilities of state-of-the-art systems, we employ knowledge distillation to specialize it for privacy-centric tasks. This model drives a dynamic three-stage pipeline: extraction, strategic sanitization, and post-hoc context restoration. Crucially, to accommodate varying user tolerances, our framework provides flexible control via \textit{Utility Priority} and \textit{Privacy Priority} modes. Experimental results show that our framework advances the privacy-utility Pareto frontier by reaching a low-leakage privacy point while preserving substantially stronger utility than existing privacy-oriented baselines.

\section{Related Work}

\paragraph{LLM Privacy-Preserving Techniques.} Extensive research addresses the risk of LLMs memorizing sensitive information from their training corpora \cite{carlini2021extracting,liu2024probing}. The most direct approach is data sanitization, which entails detecting and redacting personally identifiable information (PII) from datasets \cite{lison2021anonymisation,kandpal2022deduplicating}. While heuristic-based filtering and named entity recognition (NER) models \cite{chen2023hide} can mitigate explicit leakage, they are prone to false negatives \cite{lukas2023analyzing} and can disrupt linguistic coherence. To provide formal privacy guarantees, researchers have alternatively integrated differential privacy (DP) into the training process \cite{abadi2016deep,li2021large}. Recent works attempt to scale these methods to LLMs \cite{yu2021differentially,sinha2025vaultgemma}. However, despite its theoretical robustness, DP often incurs a substantial "utility tax", where the injection of noise inevitably compromises the model's capabilities.

Complementing these preventative strategies, machine unlearning has emerged as a post-hoc remediation technique, particularly to comply with regulations like the "Right to be Forgotten" \cite{jang2023knowledge,liu2025rethinking,liu2025learning}. The objective is to erase the influence of specific sensitive samples from a trained model without retraining. While promising, it faces the challenge of catastrophic forgetting, where the removal of specific knowledge inadvertently degrades the model's general capabilities \cite{yao2024large,xu2025obliviate}.

\paragraph{Privacy Protection in User-LLM Interactions.} In contrast to training-side defenses, inference-phase protection focuses on safeguarding user input from third-party providers. Early attempts addressed this via representation perturbation or formal privacy mechanisms. Techniques such as adding noise to input embeddings \cite{feyisetan2020privacy,plant2021cape,meehan2022sentence,du2023sanitizing} or employing differentially private decoding \cite{majmudar2022differentially,zhang2024privacyasst,zeng2025privacyrestore} aim to render inputs unreadable to humans or statistically secure. However, these methods face significant practical hurdles. Embedding-based approaches are incompatible with widespread text-only commercial APIs and remain vulnerable to inversion attacks \cite{song2020information}. Similarly, applying DP noise directly to text generation often disrupts semantic integrity, rendering prompts unintelligible to the downstream model and severely degrading task performance.

Consequently, research shifts toward direct text-level sanitization. While standard rule-based or NER methods are computationally efficient, their context-agnostic nature often necessitates the removal of task-relevant details, leading to utility collapse \cite{microsoft_2025, lison2021anonymisation}. To mitigate this, recent works leverage LLMs for more flexible protection. Approaches range from reformulating out-of-context information \cite{ngong2025protecting} to dynamically adjusting sanitization strength based on leakage risk \cite{shen2024fire}. Other strategies explore architectural or cryptographic solutions, such as delegating inference between local and remote models \cite{li2025papillon} or employing format-preserving encryption for sensitive tokens \cite{chowdhury2025prepsilonepsilonmptsanitizingsensitiveprompts}. Although superior to rigid rules, these methods often rely on static heuristics. Crucially, they overlook the combinatorial Interplay of sensitive attributes and fail to strategically adapt sanitization actions to user intent. Consequently, this lack of granularity yields suboptimal privacy-utility trade-offs.

\section{Deconstructing the Privacy-Utility Trade-off}
\label{sec:analysis}

\subsection{Data Collection and Annotation}
\label{sec:data_collection}

Rigorous empirical analysis requires a high-quality dataset consisting of user prompts annotated with their inherent sensitive statements. Our data collection pipeline integrates real-world distributions with retrieval-augmented synthesis to ensure both diversity and contextual depth.

\paragraph{Real-world and Synthetic Data.} We adopt the ShareGPT-X ~\cite{desult_2025}, LMSYS-Chat-1M ~\cite{zheng2024lmsys} and WildChat ~\cite{zhao2024wildchat} dataset to capture diverse real-world user intents. To ensure sufficient coverage of sensitive attributes, we augment this corpus with synthetic data. Specifically, we inject personal profiles from the Nemotron-Personas dataset~\cite{nemotron_personas} into user queries without sensitive information, creating samples that are both semantically coherent and rich in sensitive details.

\paragraph{Annotation and Validation.} 
We define a \textit{sensitive statement} as a discrete textual unit (e.g., a clause) that reveals specific personal attributes. We implement an automated extraction pipeline using a large reasoning model (LRM) to identify these statements. To guarantee label quality, we enforce a rigorous verification protocol. Manual inspection of a random subset ($N=200$) yields an error rate of 2.5\%, confirming the reliability of our automated pipeline. The final dataset comprises 9,757 samples (5,928 real-world and 3,829 synthetic). The supplementary material provides additional implementation details.

\subsection{Experimental Setup}
\label{sec:experimental_setup}

To empirically quantify the marginal utility of sanitization, we design a controlled experimental framework. Unlike prior black-box approaches, we structure our analysis to isolate the effects of user intent, sanitization strategy, and attribute interaction. The following subsections detail the data taxonomy and evaluation metrics operationalizing these dimensions.

\subsubsection{Data Selection and Taxonomy}

To investigate combinatorial interplay while eliminating information volume as a confounding factor, we construct a controlled subset, $\mathcal{D}_{\text{multi}}$. We filter for prompts containing a fixed number of sensitive statements (specifically, $N=5$), yielding 384 high-density samples. This standardization ensures that observed utility variations are driven by intent and content semantics rather than the quantity of disclosed attributes.

\begin{figure*}
    \centering

    \begin{subfigure}[b]{0.271\linewidth}
        \includegraphics[width=\linewidth]{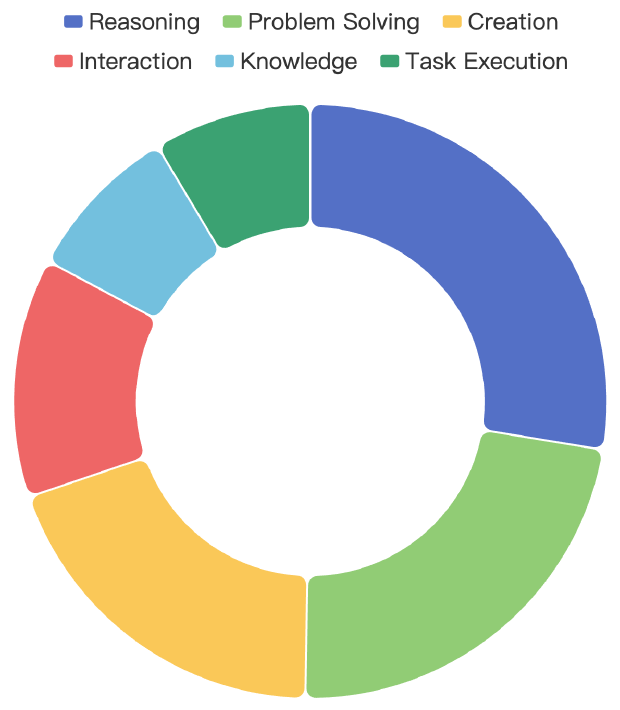}
        \caption{Intent Type}
    \end{subfigure}
    \hfill
    \begin{subfigure}[b]{0.27\linewidth}
        \includegraphics[width=\linewidth]{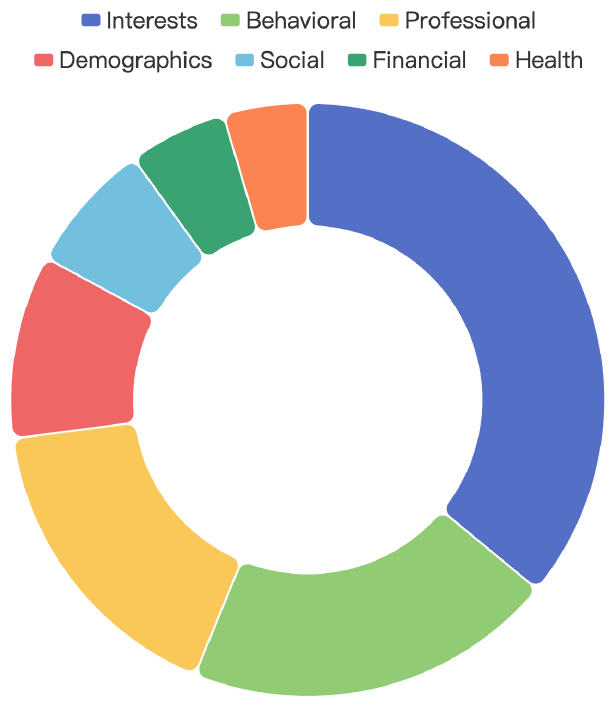}
        \caption{Privacy Type}
    \end{subfigure}
    \hfill
    \begin{subfigure}[b]{0.39\linewidth}
        \includegraphics[width=\linewidth]{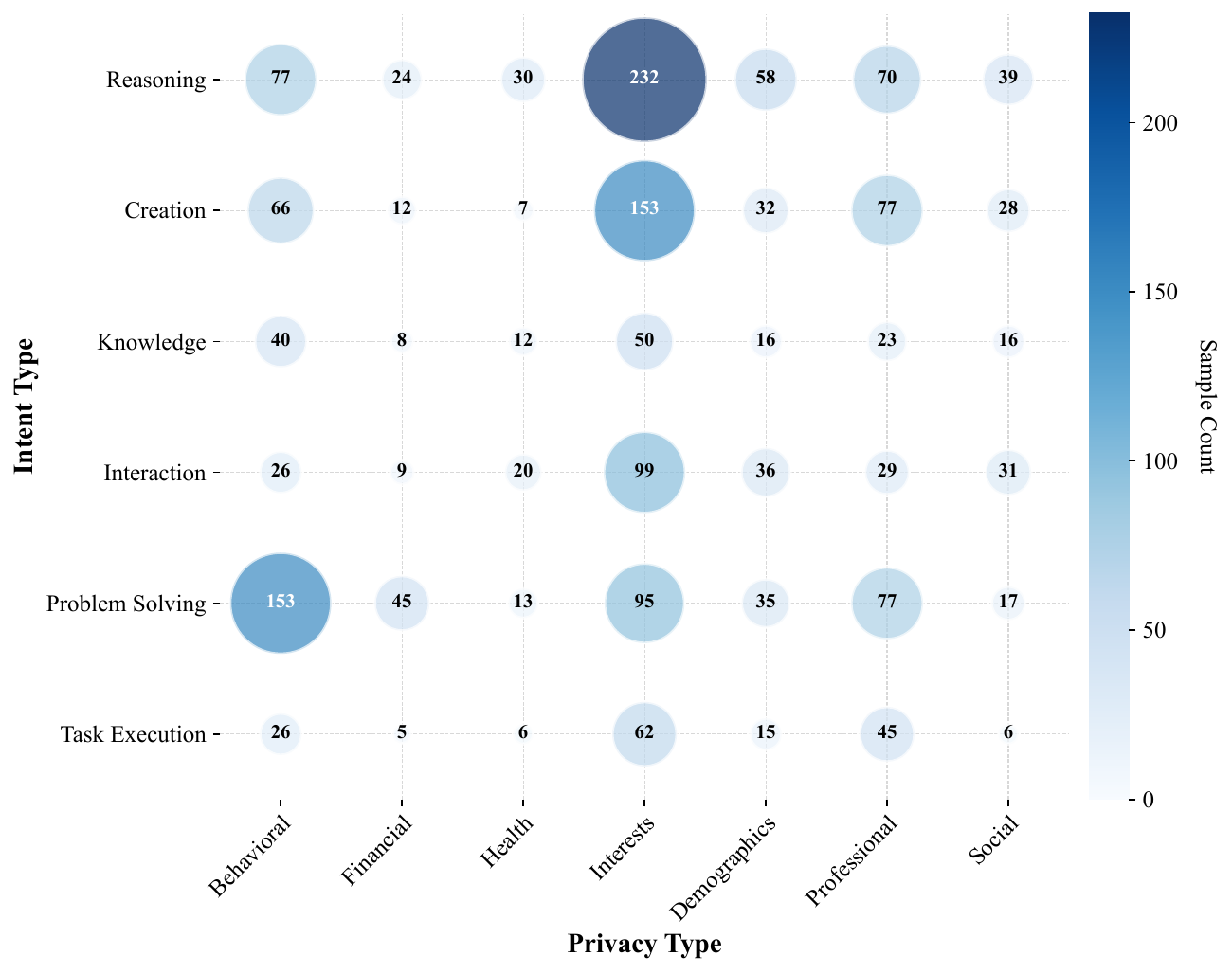}
        \caption{Joint Distribution}
    \end{subfigure}

    \caption{Dataset statistics illustrating the marginal and joint distributions of intent types and privacy types.}
    \label{fig:data_distribution}
\end{figure*}

We employ \texttt{gemini-2.5-flash}~\cite{comanici2025gemini} to annotate this subset along two orthogonal dimensions. As detailed in the supplementary material, we categorize prompts into six \textbf{Intent Types}, ranging from open-ended \textit{Creation \& Ideation} to constrained \textit{Task Execution}. Simultaneously, we classify sensitive statements into seven \textbf{Privacy Types} aligned with standard PII definitions. Figure~\ref{fig:data_distribution} visualizes the distribution across these categories.

\subsubsection{Evaluation Pipeline}

We design a pipeline to quantify the marginal utility shift ($\Delta U$) resulting from privacy sanitization. The pipeline consists of two core components: a response generator and a utility evaluator.

\paragraph{Remote Model.} We adopt \texttt{gemini-2.5-flash} \cite{comanici2025gemini} as the remote model to generate responses for original and sanitized prompts. To ensure reproducibility, we configure the decoding parameters to deterministic values with \texttt{temperature=0} and \texttt{max\_tokens=8192}.

\paragraph{Utility Metric.} We leverage \texttt{Skywork-Reward-V2-} \texttt{Llama-3.1-8B}~\cite{liu2025skywork} as the proxy for utility evaluation. Given a prompt and a response, the model assigns a scalar quality score $R$. We quantify the utility impact as the difference in reward scores between the sanitized response and the original response.

\subsection{Context-Dependent Utility: The Interplay of User Intent and Sensitive Information}

\begin{figure}[!t]
    \centering
    \includegraphics[width=\linewidth]{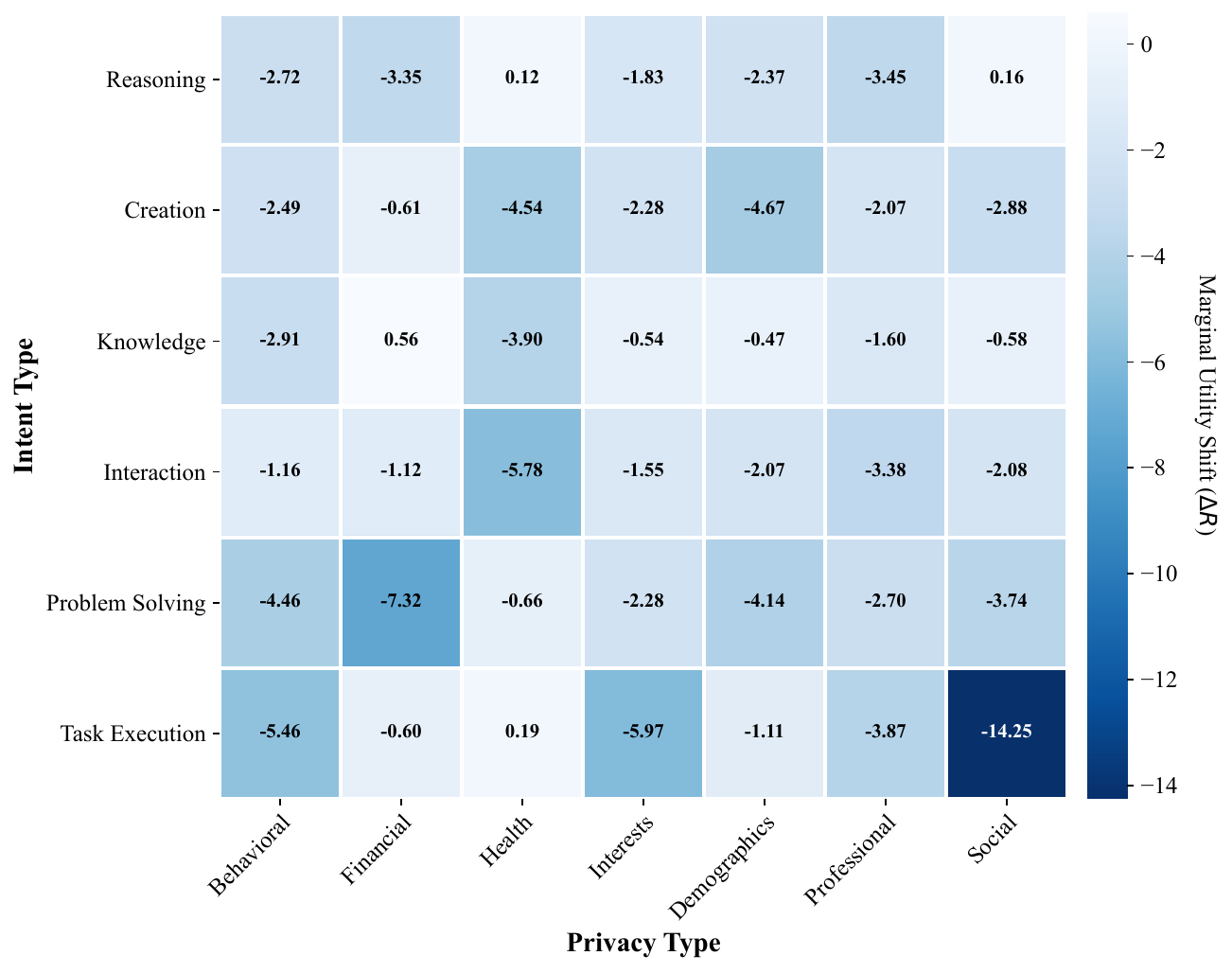}
    \caption{The utility sensitivity matrix quantifying the impact of removing sensitive information across different user intents.}
    \label{fig:intent_privacy_heatmap}
\end{figure}

We investigate how the removal of specific sensitive statements differentially impacts utility across varying user intents. To address this, we conduct a fine-grained ablation study on the subset $\mathcal{D}_{\text{multi}}$. We perform an ablation operation on each sample to remove one target statement while preserving the remaining context. We employ \texttt{gemini-2.5-flash} to execute this targeted removal following the instruction template provided in the supplementary material. We then generate responses for both original and sanitized prompts via the remote model. We quantify the utility impact as $\Delta R = R_{sanitized} - R_{original}$. A negative value indicates performance degradation. Figure~\ref{fig:intent_privacy_heatmap} visualizes the resulting utility sensitivity matrix.

Based on the empirical evidence from Figure~\ref{fig:intent_privacy_heatmap}, we summarize three findings:
\begin{itemize}[leftmargin=*]
    \item \textbf{Finding 1: Intent-Dictated Utility Sensitivity.}
    User intents strictly dictate the functional value of sensitive context. In constraint-driven tasks, specific details define the solution space. Consequently, removing \textit{Social and Relational Information} during \textit{Task Execution} causes severe performance degradation ($\Delta R = -14.25$). Conversely, logic-driven tasks often treat personal attributes as orthogonal noise. Notably, removing \textit{Health and Wellness} data in \textit{Task Execution} yields a utility gain ($\Delta R = +0.19$). This indicates that unrelated privacy leakage can distract the model from the core objective.
    \item \textbf{Finding 2: Attribute-Specific Functional Roles.}
    Sensitive attributes display distinct utility profiles rather than uniform importance. \textit{Social and Relational Information} exhibits high context sensitivity. It acts as a critical prerequisite for collaborative execution yet plays a negligible role in \textit{Personalized Interaction} ($\Delta R = -2.07$). In contrast, \textit{Health and Wellness} data demonstrates high specificity. It is vital for establishing empathy in \textit{Personalized Interaction} ($\Delta R = -5.78$) but serves as distractor noise in \textit{Analysis \& Reasoning} ($\Delta R = +0.12$).
    \item \textbf{Finding 3: Functional Coupling vs. Informational Redundancy.}
    The interaction between user intent and sensitive content dictates the optimal protection strategy. We observe functional coupling where privacy leakage is a prerequisite for utility. This is evident in the strong dependency of \textit{Problem Solving} on \textit{Financial Information}. In contrast, informational redundancy occurs when sensitive attributes provide no marginal value. The positive reward shift upon removing \textit{Financial Information} from \textit{Information Acquisition} ($\Delta R = +0.56$) confirms that privacy preservation in such contexts incurs no utility cost.
\end{itemize}


\begin{figure*}[!t]
    \centering
    \begin{subfigure}[b]{0.48\linewidth}
        \centering
        \includegraphics[width=\linewidth]{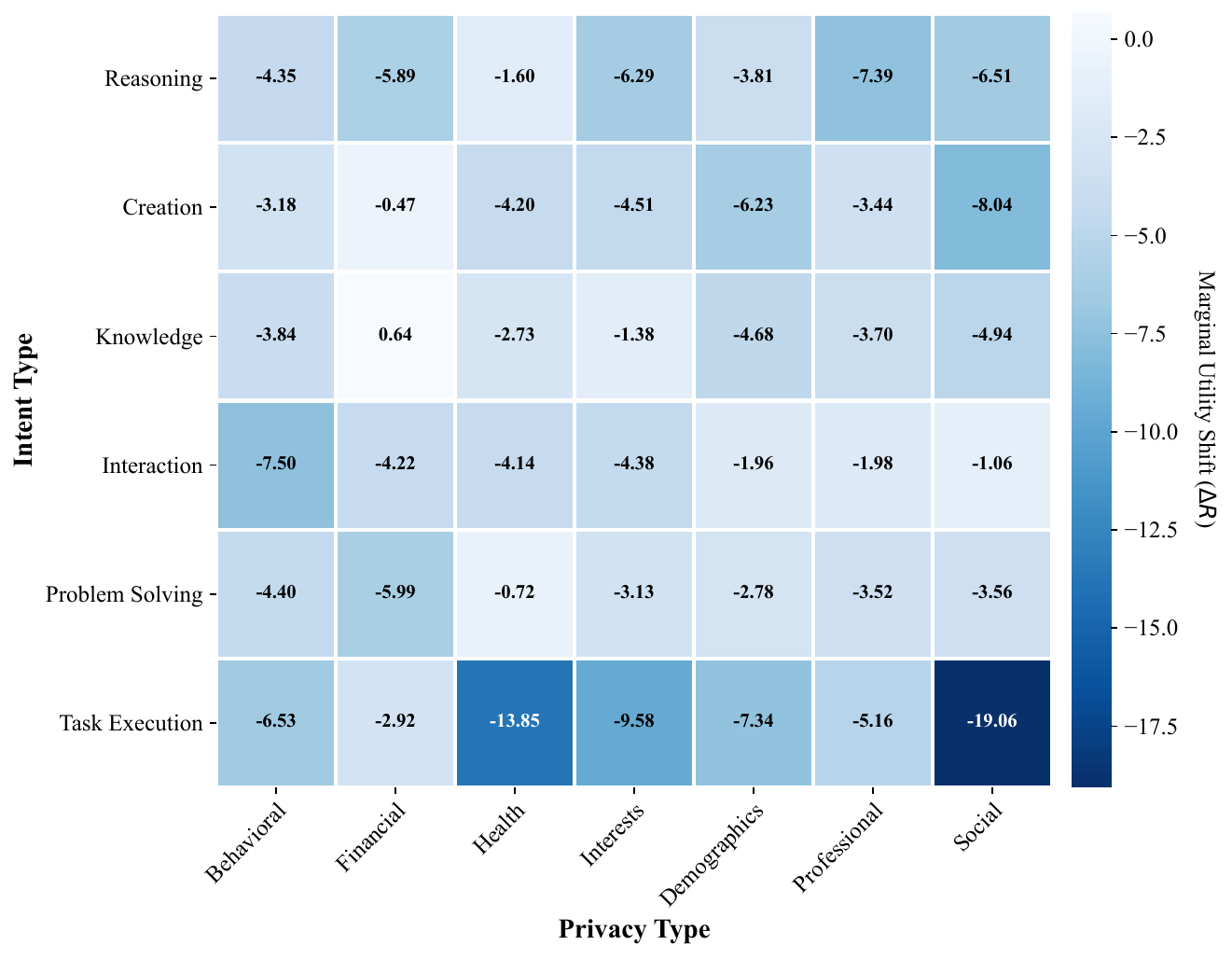}
        \caption{Impact of Replacement Strategy}
        \label{fig:replacement_impact}
    \end{subfigure}
    \hfill
    \begin{subfigure}[b]{0.48\linewidth}
        \centering
        \includegraphics[width=\linewidth]{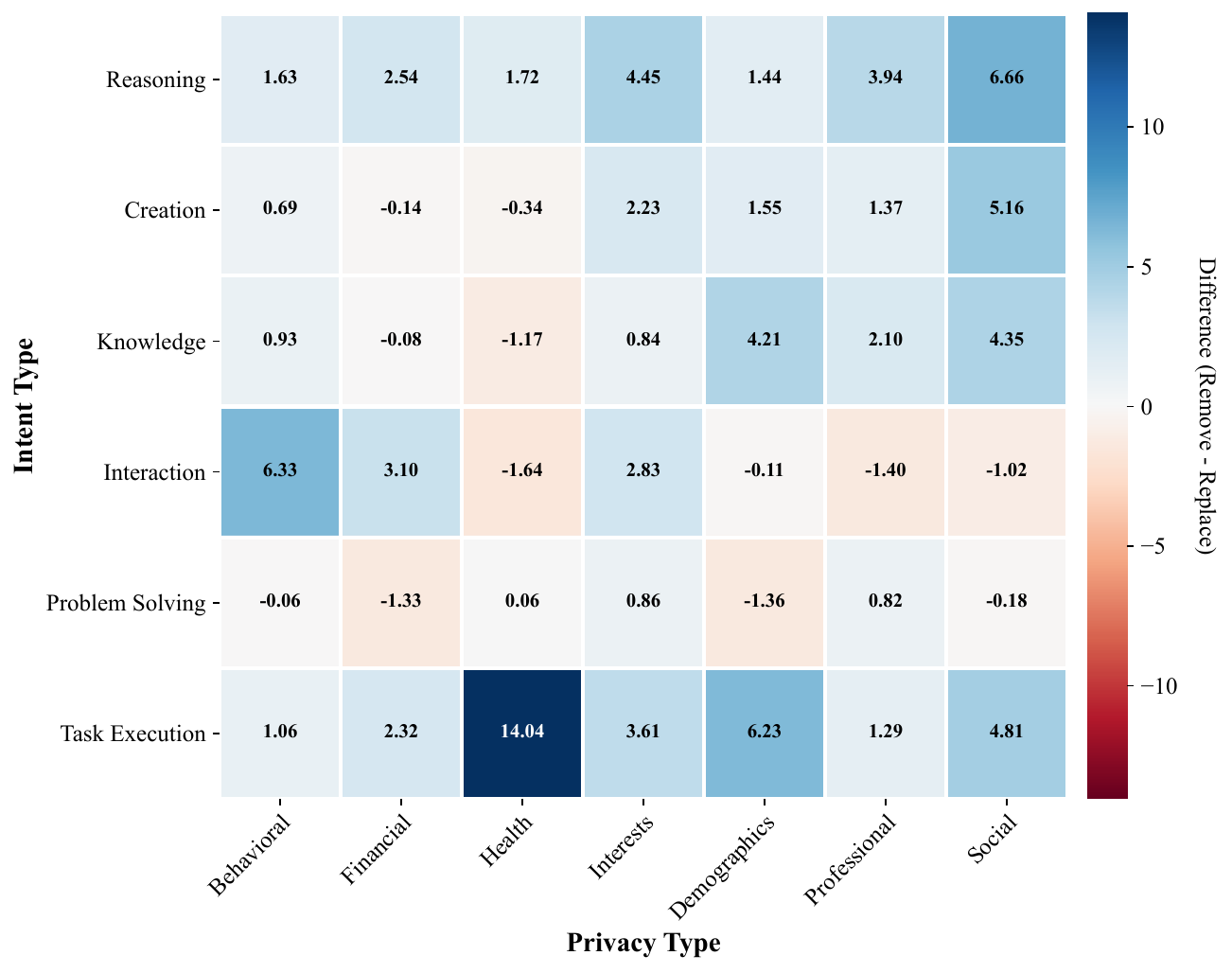}
        \caption{Strategic Differential}
        \label{fig:strategy_diff}
    \end{subfigure}
    
    \caption{Comparative utility analysis of sanitization strategies: (a) Quantifies the marginal utility shift caused by substituting sensitive attributes with synthetic placeholders. (b) Visualizes the strategic differential, where positive values (blue) indicate removal preserves more utility, while negative values (red) favor replacement.}
    \label{fig:sanitization_analysis_combined}
\end{figure*}

\subsection{Strategic Adaptation: The Efficacy of Removal versus Replacement}
\label{sec:strategic_coupling}

Removal serves as the standard baseline for sanitization. However, it often fractures the semantic coherence of the prompt. To address this, we evaluate a \textit{Replacement} strategy, which substitutes sensitive attributes with synthetic placeholders using the supplementary template, and visualize the resulting utility impact in Figure~\ref{fig:replacement_impact}. To rigorously compare the relative efficacy of these two approaches, we define the strategic differential as $D = \Delta R_{\text{remove}} - \Delta R_{\text{replace}}$. The distribution of this differential is presented in Figure~\ref{fig:strategy_diff}, where positive values (blue regions) indicate that removal preserves more utility, while negative values (red regions) favor replacement.

Our comparative analysis reveals two distinct patterns driven by the functional role of the information:
\begin{itemize}[leftmargin=*]
    \item \textbf{Finding 1: Factual Integrity and the Risk of False Premises.} 
    For tasks grounded in objective execution, accuracy is paramount. In these scenarios, providing a fabricated value via replacement is significantly more damaging than omitting the data through removal. We observe that synthetic placeholders often function as false premises, causing the model to hallucinate solutions based on incorrect pathologies or constraints. This is evident in the interaction between \textit{Task Execution} and \textit{Health Information}, where the strategic differential peaks at $D = +14.04$. Consequently, for logic-driven tasks, silence is superior to noise.
    \item \textbf{Finding 2: Structural Coherence and Narrative Anchoring.} 
    Conversely, replacement strategy demonstrates a comparative advantage when the user seeks structural guidance or empathy rather than factual precision. In these contexts, sensitive statements act as conversational anchors that maintain the dialogue flow. For instance, in \textit{Personalized Interaction}, replacing \textit{Health} data allows the model to preserve an empathetic tone, thereby outperforming removal ($D = -1.64$). Similarly, in \textit{Problem Solving}, using a dummy \textit{Financial} figure ($D = -1.33$) enables the model to demonstrate a correct calculation process. In such cases, the structural validity of the response outweighs the factual accuracy of the input.
\end{itemize}


\subsection{Combinatorial Interplay: The Synergy and Antagonism of Sensitive Information}
\label{sec:interaction_effects}

\begin{figure}[t]
    \centering
    \includegraphics[width=\linewidth]{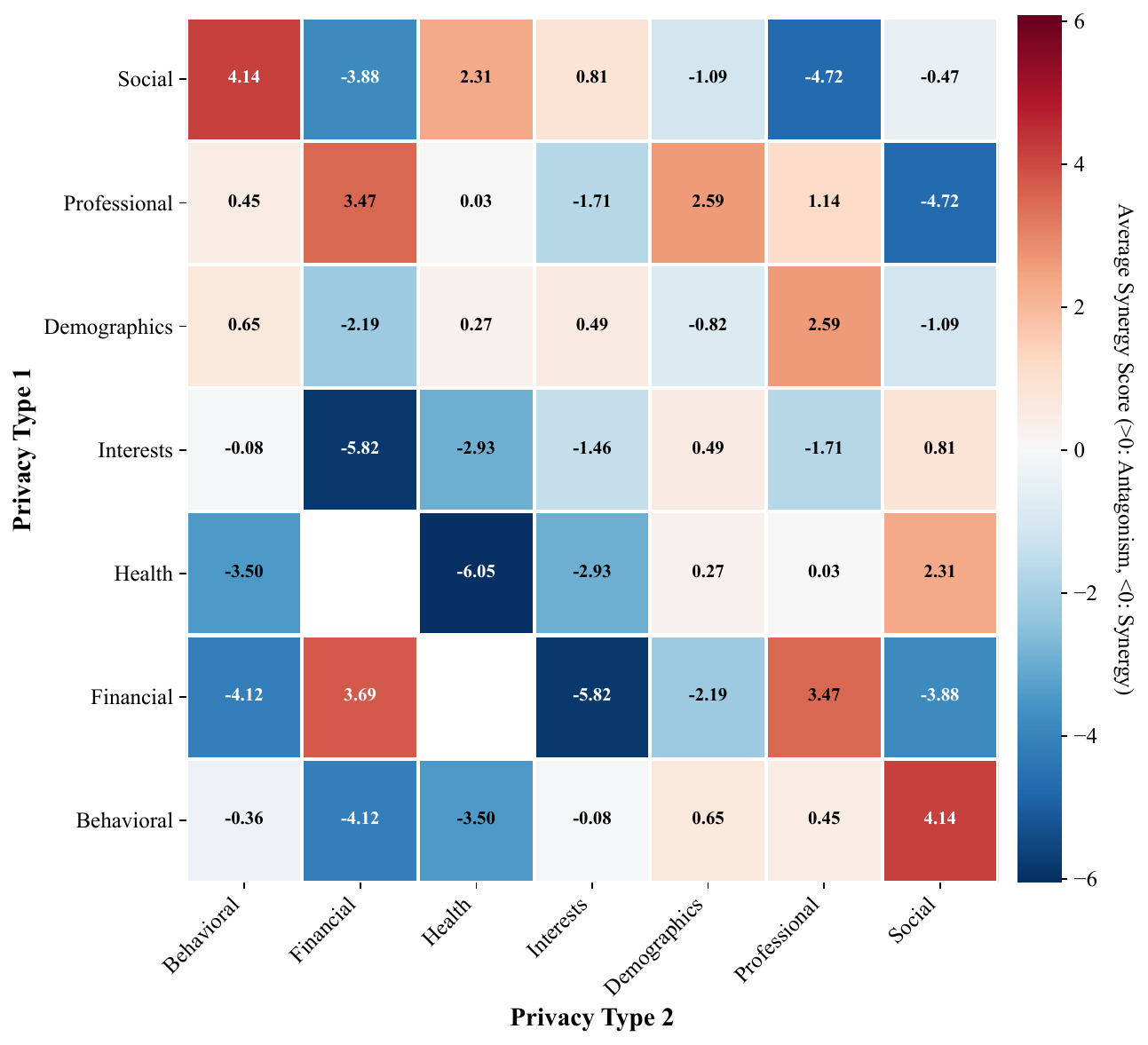}
    \caption{The global interaction matrix illustrating the combinatorial interplay of sensitive statements. Blank cells indicate no available data for that privacy type pair.}
    \label{fig:interaction_overall_heatmap}
\end{figure}

Sensitive statements rarely function in isolation; rather, they work in an interconnected manner where one attribute affects the utility of another. To measure these interactions, we conduct pairwise removals on $\mathcal{D}_{\text{multi}}$ using the prompt template provided in the supplementary material. For every pair of sensitive statements $(A, B)$, we remove both simultaneously and measure the utility shift $\Delta R(A, B)$. We define the interaction score ($I_{A,B}$) as the difference between the joint impact and the sum of individual impacts:
\begin{equation}
    I_{A,B} = \Delta R(A, B) - (\Delta R(A) + \Delta R(B))
\end{equation}
A negative score ($I < 0$) indicates synergy, where removing both attributes together causes a utility drop significantly larger than the sum of removing them individually. Conversely, a positive score ($I > 0$) indicates antagonism, implying that the attributes are redundant and provide overlapping information. We visualize the global distribution of these effects in Figure~\ref{fig:interaction_overall_heatmap}; the supplementary material provides the user-intent breakdown.

Our analysis yields three findings:
\begin{itemize}[leftmargin=*]
    \item \textbf{Finding 1: Narrative Cohesion Drives Synergy.} 
    Strong synergistic effects emerge when two statements are logically interlinked to construct a coherent user story. As evidenced by the deep blue regions in the global heatmap, we observe significant synergy between \textit{Health and Wellness} and \textit{Interests, Beliefs, and Opinions} ($I = -6.05$). These pairs often establish a causal link, such as a health condition motivating a specific lifestyle change. Consequently, severing both links destroys the causal chain, leaving the model with no basis to infer the user's underlying motivation.
    \item \textbf{Finding 2: Informational Redundancy Drives Antagonism.} 
    Conversely, antagonistic effects appear when statements provide overlapping signals. In the global view, \textit{Behavioral Data} and \textit{Social Information} exhibit high antagonism ($I = +4.14$). For example, a social status like "Student" inherently implies behavioral patterns such as "Studying," rendering the explicit statement of behavior redundant. In this case, removing only one attribute fails to protect privacy, as the model can reconstruct the missing context from its counterpart.
    \item \textbf{Finding 3: User Intent Dictates Interaction Patterns.} 
    The nature of interaction is strictly determined by user intent rather than being inherent to the data categories. In \textit{Task Execution}, we observe extreme antagonism where eliminating both \textit{Social Information} and \textit{Interests} yields a peak score of $I = +47.8$. This confirms that simultaneously removing unrelated noise significantly amplifies model focus. Distinctly, \textit{Personalized Support} demonstrates a selective anchoring effect, where the presence of critical \textit{Health} context renders \textit{Professional Background} functionally redundant ($I = +28.2$).
\end{itemize}


\section{Framework Implementation and Evaluation}
\label{sec:framework_eval}

Section~\ref{sec:analysis} established three key insights regarding the privacy-utility trade-off: context-dependent utility, strategic adaptation, and combinatorial interplay. Existing methods relying on static rules cannot address these complexities. In this section, we propose an Intent-Driven Local Protection Framework that directly applies these insights. Instead of using rigid heuristics, we formulate our empirical findings into explicit reasoning instructions. Through knowledge distillation, we embed this logic into a lightweight local model, teaching it to evaluate data sensitivity based on user intent. This enables the framework to execute a precise three-stage pipeline: Extraction, Sanitization, and Restoration.

\subsection{Architecture Design}

The inference workflow of our framework consists of three sequential modules.

\paragraph{Stage 1: Sensitive Information Extraction.}
Given a user prompt $P$, the local model first identifies and extracts the set of sensitive statements $\mathcal{S} = \{s_1, s_2, \dots, s_n\}$. This process requires precise entity recognition and contextual understanding to capture both explicit PII and implicit attribute disclosure.

\paragraph{Stage 2: Strategic Planning and Sanitization.}
This module constitutes the core decision-making engine. Guided by the principles derived in Section~\ref{sec:analysis}, the model evaluates each statement $s_i \in \mathcal{S}$ based on the user's intent and the privacy category. It generates a protection plan $\pi$ that assigns an action $a_i \in \{\textsc{Keep}, \textsc{Remove}, \textsc{Replace}\}$ to each statement.

Simultaneously, the model executes this plan to transform the original prompt $P$ into a sanitized version $P'$.
\begin{equation}
    P', \pi = \mathcal{M}_{\text{local}}(P, \mathcal{S})
\end{equation}
Crucially, the decision logic accounts for the interaction effects between statements. The model removes antagonistic pairs to prevent redundancy leakage and preserves synergistic pairs when necessary for task utility.

\paragraph{Stage 3: Response Restoration.}
The sanitized prompt $P'$ is transmitted to the untrusted remote model, which returns a generic response $R'$. To ensure the final output remains personalized and relevant, our local model performs a post-processing step. It utilizes the stored plan $\pi$ and the original sensitive statements $\mathcal{S}$ to re-inject necessary context into $R'$.
\begin{equation}
    R_{\text{final}} = \mathcal{M}_{\text{local}}(R', \pi, \mathcal{S})
\end{equation}
This ensures that the user receives a high-utility response without ever exposing raw sensitive data to the remote server.

\subsection{Model Distillation}

To enable a lightweight local model to perform these complex reasoning tasks, we employ knowledge distillation. We leverage \texttt{Deepseek-V4-Flash}~\cite{xu2026deepseek} as the teacher to synthesize high-quality training data, and fine-tune a compact \texttt{Qwen3-4B} \cite{qwen3technicalreport} student model. We refer to the resulting privacy-specialized student model as \texttt{Veilmind-4B}. The training data synthesis covers three aspects:

\paragraph{Synthesis 1: Reasoning-Enhanced Extraction.}
Although our dataset from Section~\ref{sec:data_collection} already contains prompt-statement pairs $(P, \mathcal{S})$, direct supervision is insufficient for handling ambiguous cases. We prompt the teacher model to generate a detailed reasoning chain that explains why specific segments are classified as sensitive. This helps the student model learn the criteria for sensitivity detection. The prompt template is provided in the supplementary material.

\paragraph{Synthesis 2: Sanitization.}
Users exhibit varying tolerances for the privacy-utility trade-off. To accommodate this, we introduce two distinct operating modes:
\begin{itemize}[leftmargin=*]
    \item \textbf{Utility Priority Mode:} The model prioritizes task performance. It retains sensitive information if it serves as a critical constraint (e.g., financial data in problem-solving). The corresponding prompt template is provided in the supplementary material.
    \item \textbf{Privacy Priority Mode:} The model minimizes leakage. It aggressively sanitizes information unless it renders the prompt incoherent. The corresponding prompt template is provided in the supplementary material.
\end{itemize}
We instruct the teacher model to simulate both perspectives. For each prompt, we generate reasoning traces and revised prompts under both modes. This exposes the student model to the decision boundaries of different protection strategies.

\paragraph{Synthesis 3: Restoration.}
Finally, we synthesize the restoration phase. We feed the teacher model the sanitized prompt $P'$, the execution plan $\pi$, and the remote response $R'$. The teacher generates a restored response $R_{\text{final}}$ that seamlessly integrates the withheld information. The restoration template is provided in the supplementary material.

\subsection{Evaluation}

We evaluate our framework on Uprise, the manually verified subset of our dataset Uprise ($N=200$) described in Section~\ref{sec:data_collection}. We additionally evaluate our framework on Pupa-tnb~\cite{li2025papillon}, a benchmark of real-world user queries containing personally identifiable information (PII), the results of which are provided in the supplementary material.

\paragraph{Baselines.} We benchmark against two representative methods. \textbf{Papillon}~\cite{li2025papillon} employs a multi-stage delegation framework where a local proxy synthesizes sanitized queries and aggregates responses. We evaluate its zero-shot \textit{Base} variant and the DSPy-tuned \textit{Optimized} variant~\cite{khattab2023dspy}. \textbf{PUFT}~\cite{ngong2025protecting} utilizes Contextual Integrity to reformulate prompts by retaining only task-essential details. We examine both its \textit{Static} variant relying on predefined attribute lists and the \textit{Dynamic} variant that adapts to specific interaction contexts.

\begin{table*}[t]
\centering
\small
\setlength{\tabcolsep}{3.5pt}
\begin{tabular*}{\textwidth}{@{\extracolsep{\fill}}l c c c c c c c c@{}}
\toprule
\textbf{Model} &
\multirow{2}{*}{\textbf{\shortstack{Extraction\\ Success Rate}}} &
\multicolumn{3}{c}{\textbf{Privacy-Mode Sanitization}} &
\multicolumn{3}{c}{\textbf{Utility-Mode Sanitization}} &
\multirow{2}{*}{\textbf{\shortstack{Restoration\\Utility Gain}}} \\
\cmidrule(lr){3-5}\cmidrule(lr){6-8}
& & \textbf{Remove} & \textbf{Replace} & \textbf{Keep}
& \textbf{Remove} & \textbf{Replace} & \textbf{Keep} \\
\midrule
Qwen3-4B & 52.4 & 54.5 & 37.4 & 8.1 & 34.7 & 40.8 & 24.5 & +3.5 \\
Veilmind-4B & 72.7 & 33.6 & 48.6 & 17.9 & 11.8 & 25.4 & 62.8 & +7.0 \\
DeepSeek-V4-Flash & 77.4 & 41.0 & 36.3 & 22.8 & 10.0 & 27.6 & 62.5 & +10.0 \\
\bottomrule
\end{tabular*}
\caption{Strategy-level effect of fine-tuning on privacy extraction, sanitization action, and restoration. All values are percentages.}
\label{tab:overall-comparison}
\end{table*}

\begin{figure}[t]
    \centering
    \includegraphics[width=\linewidth]{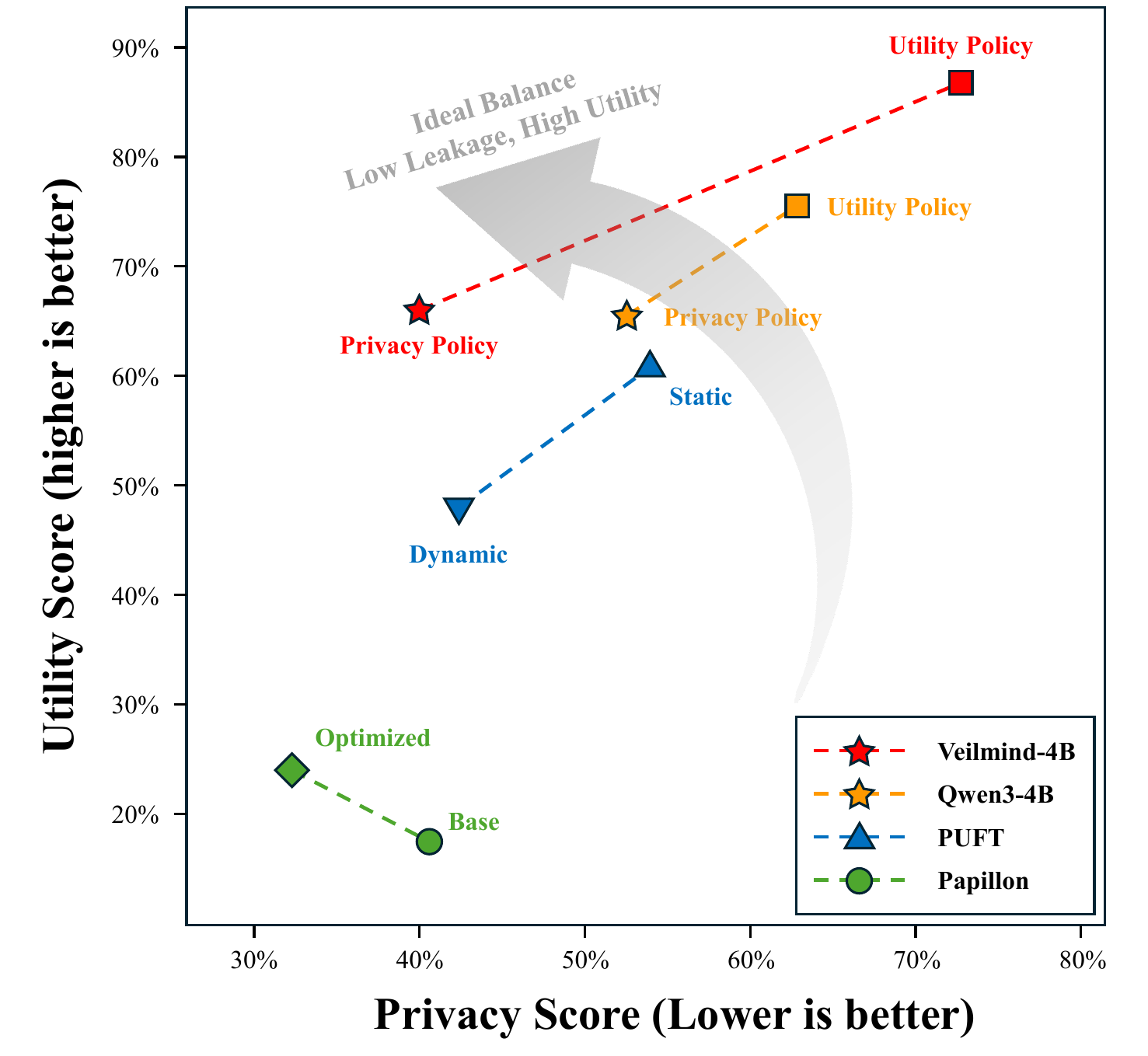}
    \caption{The privacy-utility trade-off comparison on Uprise.}
    \label{fig:main_results}
\end{figure}

\paragraph{Metrics.} We employ \texttt{Deepseek-V4-Flash}~\cite{xu2026deepseek} as an impartial judge to compute two core metrics. For \textbf{Utility Score}, we adopt a pairwise comparison approach, reporting the win/tie rate where the sanitized response is deemed comparable to or better than the original. For \textbf{Privacy Score}, we quantify leakage by calculating the retention rate of sensitive statements in the revised prompt, where a lower score indicates stronger protection.

\paragraph{Main Results.} Figure~\ref{fig:main_results} illustrates the superior privacy-utility trade-off achieved by our framework. In \textbf{Privacy Priority Mode}, our method achieves a significantly higher utility score while maintaining only a slightly higher privacy score than Papillon Optimized, and outperforms other baselines in both privacy and utility, validating the efficacy of adaptive sanitization driven by user intent. Conversely, the \textbf{Utility Priority Mode} preserves critical constraints to yield exceptional response quality. These results confirm that our framework not only advances the Pareto frontier but also provides users with flexible control to balance protection requirements against task performance.

\paragraph{Impact of Model Distillation.}
To assess the necessity of fine-tuning, we directly apply our three-stage framework to the base \texttt{Qwen3-4B} model. The main results show that the base model is weaker than \texttt{Veilmind-4B}, but still outperforms PUFT and Papillon, indicating that the extraction-sanitization-restoration design is effective even before distillation. Fine-tuning further turns this strong initialization into a state-of-the-art privacy-utility trade-off: on Uprise, utility-mode utility improves from 75.5\% to 86.8\%, while privacy-mode leakage decreases from 52.6\% to 40.5\%. Table~\ref{tab:overall-comparison} explains this gain at the strategy level. First, privacy extraction coverage increases from 52.4\% to 72.7\%, approaching \texttt{DeepSeek-V4-Flash} model at 77.4\%. Second, fine-tuning makes sanitization less deletion-heavy and more discriminative. In Privacy Priority Mode, the removal rate drops from 54.5\% to 33.6\%, while replacement becomes the dominant action at 48.6\%; in Utility Priority Mode, the keep rate rises from 24.5\% to 62.8\%, preserving more low-risk information that supports answer quality. Third, restoration also improves, with the utility gain increasing from +3.5\% to +7.0\%. These changes suggest that fine-tuning transfers the handling strategy of \texttt{DeepSeek-V4-Flash} to the \texttt{Qwen3-4B} model, yielding stronger extraction, more balanced sanitization, and more effective utility recovery.


\begin{table}[t]
\centering
\small
\setlength{\tabcolsep}{1.5pt}
\begin{tabular*}{\linewidth}{@{\extracolsep{\fill}}l c c c@{}}
\toprule
\textbf{Model} &
\textbf{\shortstack{No Extraction}} &
\textbf{Privacy $\downarrow$} &
\textbf{Utility $\uparrow$} \\
\midrule
w/o Extract &
81 &
\shortstack{71.6\% (+19.0)} &
\shortstack{82.00\% (+16.5)} \\
w/o Sani. Guide &
26 &
\shortstack{60.7\% (+8.1)} &
\shortstack{71.5\% (+6.0)} \\
w/o Restore &
30 &
\shortstack{52.6\% (+0.0)} &
\shortstack{62.0\% (-3.5)} \\
\midrule
Qwen3-4B &
30 &
52.6\% &
65.5\% \\
\bottomrule
\end{tabular*}
\caption{Ablation study of the three-stage framework. No Extraction counts prompts for which the model outputs no extracted privacy item. }
\label{tab:stage-ablation}
\end{table}

\paragraph{Stage Ablation.}
Table~\ref{tab:stage-ablation} validates the necessity of the Extraction, Sanitization, and Restore stages on \texttt{Qwen3-4B} model. Sanitization without Extraction causes the largest privacy degradation: leakage increases from 52.6\% to 71.6\%, while the number of prompts with no privacy extracted rises from 30 to 81. The higher utility of this variant therefore reflects insufficient privacy identification before Sanitization rather than a better privacy-utility trade-off. Removing the Sanitization Guideline also weakens protection, increasing leakage to 60.7\%, which shows that extracted privacy items must be converted into reliable removal or replacement decisions. Finally, removing Restoration leaves privacy unchanged but reduces utility to 62.0\%, isolating the role of Restore in recovering answer quality after sanitization. Together, these ablations show that Extraction provides privacy coverage, Sanitization controls leakage through concrete edit decisions, and Restore recovers utility without increasing privacy exposure.

\section{Conclusion}

In this paper, we investigate the trade-off between privacy and utility in LLM interactions. Our analysis indicates that the marginal utility of sensitive information is highly context-dependent. It relies on specific user intents and the combinatorial interplay of privacy attributes. We demonstrate that optimal sanitization requires adaptive strategies. Guided by these insights, we propose a local framework that dynamically adjusts sanitization strategies via a distill-and-deploy pipeline. Experiments demonstrate that our approach effectively balances privacy protection with response quality compared to static baselines. Future work will extend this intent-centric paradigm to multi-turn interactions, where user goals and privacy boundaries evolve progressively across the conversation history.





\bibliography{aaai2027}

\appendix

\section{Data Collection Details}
\label{app:data_details}

This appendix provides the implementation details for the data collection and annotation pipeline.

\subsection{Real-world Data Filtering}
We sourced real-world prompts from three public dialogue corpora: ShareGPT-X~\cite{desult_2025}, LMSYS-Chat-1M~\cite{zheng2024lmsys}, and WildChat~\cite{zhao2024wildchat}. The data synthesis code normalizes the user-turn fields across these sources and filters the resulting prompt pool by language, length, and duplication. Specifically, we applied the following criteria:
\begin{itemize}
\item \textbf{Language:} Retained prompts whose dataset-level language tags fall into the configured multilingual allowlist. During balanced collection, prompts are grouped into English, Chinese, and Other with a target ratio of 60:20:20.
\item \textbf{Length constraints:} Restricted the token length $L$ of user prompts to the range $16 < L < 4096$, and bucketed the retained prompts into four ranges: $L<64$, $64\le L<256$, $256\le L<1024$, and $L\ge1024$.
\item \textbf{Quota-aware balancing:} Deduplicated exact prompt text and then balanced sampling by source, language group, length bucket, and domain bucket. When a language-domain bucket exceeded its quota, we applied stride-based weighted subsampling rather than simply truncating the earliest examples, which limits overrepresented buckets while preserving diversity.
\end{itemize}
This filtering process yielded an initial corpus of 61,275 candidate prompts.

\subsection{Synthetic Data Generation Pipeline}
To enhance the diversity and scale of the dataset, we employ a multi-stage injection pipeline:

\paragraph{Persona Retrieval.} We utilize \texttt{Qwen3-Embedding-}\texttt{8B} \cite{qwen3embedding} to encode both candidate prompts and personas from the Nemotron-Personas dataset \cite{nemotron_personas}. For each prompt, we retrieve the top-8 candidate personas based on cosine similarity and re-rank them using \texttt{Qwen3-Reranker-}\texttt{8B} \cite{qwen3embedding} to select the single most contextually relevant persona.

\paragraph{Suitability Assessment.} Not all prompts are suitable for privacy injection. We utilize \texttt{Deepseek-V4-} \texttt{Flash}~\cite{xu2026deepseek} to score the injection suitability of each prompt on a scale of 1 to 5, employing the instruction template provided in Table~\ref{tab:suitability_prompt}. Only candidates with a score $\ge 3$ were selected for processing.

\paragraph{Injection and Extraction.} The selected persona was injected into the prompt using the instruction template shown in Table~\ref{tab:injection_prompt}. Subsequently, we employed \texttt{Deepseek-} \texttt{V4-Flash}~\cite{xu2026deepseek} to extract the lists of sensitive statements $\mathcal{I}$ from both real and synthetic prompts, following the extraction criteria detailed in Table~\ref{tab:extraction_prompt}.

\subsection{Quality Assurance}
To ensure quality, we implemented an double-check process where the model re-evaluates its extracted sensitive statements using the prompt presented in Table~\ref{tab:verification_prompt}. From the final corpus, we reserved 200 manually validated samples as the benchmark for final evaluation, while the remaining samples are used for the empirical study.

\section{Taxonomy Definitions}
\label{app:taxonomy_definitions}

Table~\ref{tab:taxonomy_definitions} summarizes the taxonomy definitions used in our analysis. We categorize user requests into six intent types and sensitive attributes into seven privacy types. These categories provide the shared vocabulary for the distribution analysis, interaction analysis, and qualitative case studies.

\section{Experiment Details and Additional Results}
\label{app:experiment_details}

\begin{table}[!t]
    \centering
    \small
    \begin{tabular*}{0.82\linewidth}{@{\extracolsep{\fill}}lrr}
        \toprule
        \textbf{Language} & \textbf{Count} & \textbf{Ratio} \\
        \midrule
        English & 6,232 & 63.87\% \\
        Chinese & 1,691 & 17.33\% \\
        Portuguese & 367 & 3.76\% \\
        Spanish & 247 & 2.53\% \\
        Russian & 234 & 2.40\% \\
        French & 213 & 2.18\% \\
        Other & 773 & 7.92\% \\
        \midrule
        Total & 9,757 & 100.00\% \\
        \bottomrule
    \end{tabular*}
    \caption{Language composition of the prompt data used for distillation. All languages outside the top six are grouped as Other.}
    \label{tab:extraction_language_distribution}
\end{table}

\begin{table}[!t]
    \centering
    \small
    \begin{tabular*}{0.86\linewidth}{@{\extracolsep{\fill}}lrr}
        \toprule
        \textbf{Intent} & \textbf{Count} & \textbf{Ratio} \\
        \midrule
        Creation & 2,141 & 21.94\% \\
        Problem Solving & 2,005 & 20.55\% \\
        Knowledge & 1,922 & 19.70\% \\
        Task Execution & 1,716 & 17.59\% \\
        Reasoning & 1,083 & 11.10\% \\
        Interaction & 890 & 9.12\% \\
        \midrule
        Total & 9,757 & 100.00\% \\
        \bottomrule
    \end{tabular*}
    \caption{Intent composition of the prompt data used for distillation.}
    \label{tab:extraction_intent_distribution}
\end{table}

\begin{table}[!t]
    \centering
    \small
    \begin{tabular*}{0.86\linewidth}{@{\extracolsep{\fill}}lrr}
        \toprule
        \textbf{Privacy Type} & \textbf{Count} & \textbf{Ratio} \\
        \midrule
        Interests & 14,043 & 26.21\% \\
        Professional & 13,083 & 24.42\% \\
        Demographics & 10,469 & 19.54\% \\
        Behavioral & 9,939 & 18.55\% \\
        Social & 3,106 & 5.80\% \\
        Health & 1,666 & 3.11\% \\
        Financial & 1,266 & 2.36\% \\
        \midrule
        Total & 53,572 & 100.00\% \\
        \bottomrule
    \end{tabular*}
    \caption{Privacy-type composition of the prompt data used for distillation. Ratios are computed over extracted claims.}
    \label{tab:extraction_privacy_type_distribution}
\end{table}

\subsection{Privacy-Attribute Interactions Across Intents}
\begin{figure*}[t]
    \centering
    \includegraphics[width=0.9\linewidth]{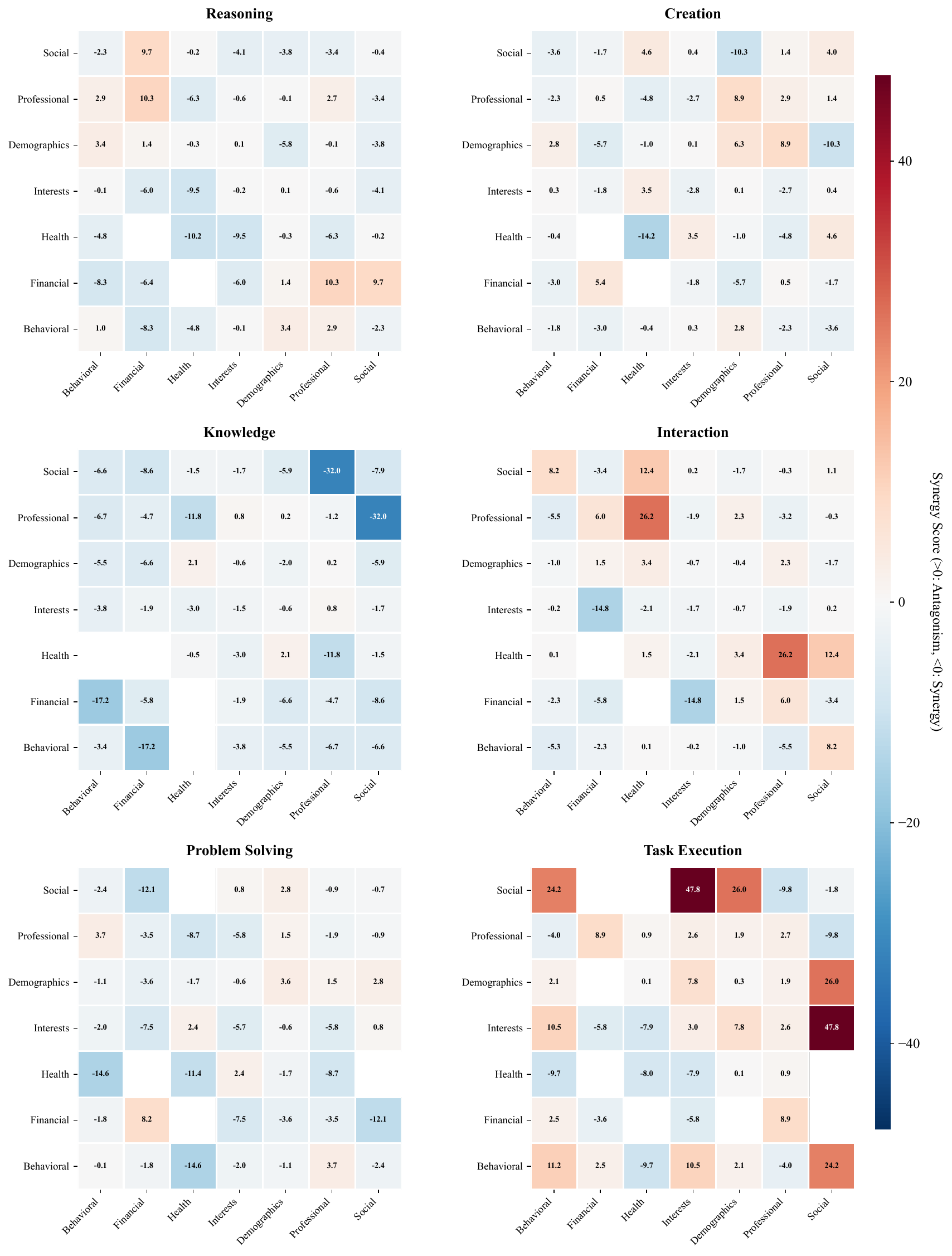}
    \caption{Disaggregated interaction heatmaps revealing how user intent modulates combinatorial interplay. The nature of interaction shifts dramatically across tasks.}
    \label{fig:interaction_intent_heatmap}
\end{figure*}

Figure~\ref{fig:interaction_intent_heatmap} reports the synergy and antagonism patterns among privacy-attribute combinations under different user intents. The result shows that privacy interactions are task-dependent: the same pair of attributes can have different utility and leakage effects when the underlying intent changes.

\begin{table*}[ht]
    \centering
    \small
    \begin{tabular}{lcccc}
        \toprule
        \textbf{Method} & \multicolumn{2}{c}{\textbf{Uprise}} & \multicolumn{2}{c}{\textbf{Pupa-tnb}} \\
        \cmidrule(lr){2-3} \cmidrule(lr){4-5}
        & \textbf{Utility} ($\uparrow$) & \textbf{Privacy} ($\downarrow$) & \textbf{Utility} ($\uparrow$) & \textbf{Privacy} ($\downarrow$) \\
        \midrule
        PUFT - Static & 60.50 & 53.95 & 28.81 & 13.53 \\
        PUFT - Dynamic & 48.00 & 42.40 & 25.00 & 6.10 \\
        \midrule
        Papillon - Base & 17.50 & 40.63 & 27.12 & 10.00 \\
        Papillon - Optimized & 24.00 & 32.31 & 24.89 & 2.75 \\
        \midrule
        Qwen3-4B - Privacy Priority & 65.50 & 52.57 & 56.96 & 15.99 \\
        Qwen3-4B - Utility Priority & 75.50 & 62.86 & 64.98 & 27.14 \\
        \midrule
        \textbf{Veilmind-4B - Privacy Priority} & \textbf{66.00} & \textbf{40.50} & \textbf{58.05} & \textbf{14.76} \\
        \textbf{Veilmind-4B - Utility Priority} & \textbf{87.00} & \textbf{72.78} & \textbf{77.12} & \textbf{49.14} \\
        \bottomrule
    \end{tabular}
    \caption{Detailed numerical results for privacy-utility trade-off comparison. All values are reported as percentages.}
    \label{tab:detailed_results}
\end{table*}

\subsection{Distillation Details}
All distillation experiments were conducted on a single NVIDIA H200 GPU. We use \texttt{Qwen3-4B} as the backbone model and train \texttt{Veilmind-4B} with full-parameter supervised fine-tuning through the ms-swift framework \cite{zhao2024swiftascalablelightweightinfrastructure}. The teacher annotations used for distillation are generated by \texttt{Deepseek-V4-Flash}, while the student model learns to perform privacy extraction, sanitization, and restoration locally. Tables~\ref{tab:extraction_language_distribution}, \ref{tab:extraction_intent_distribution}, and~\ref{tab:extraction_privacy_type_distribution} summarize the language, intent, and privacy-type composition of the final prompts used for distillation.

\begin{figure}[t]
    \centering
    \includegraphics[width=\linewidth]{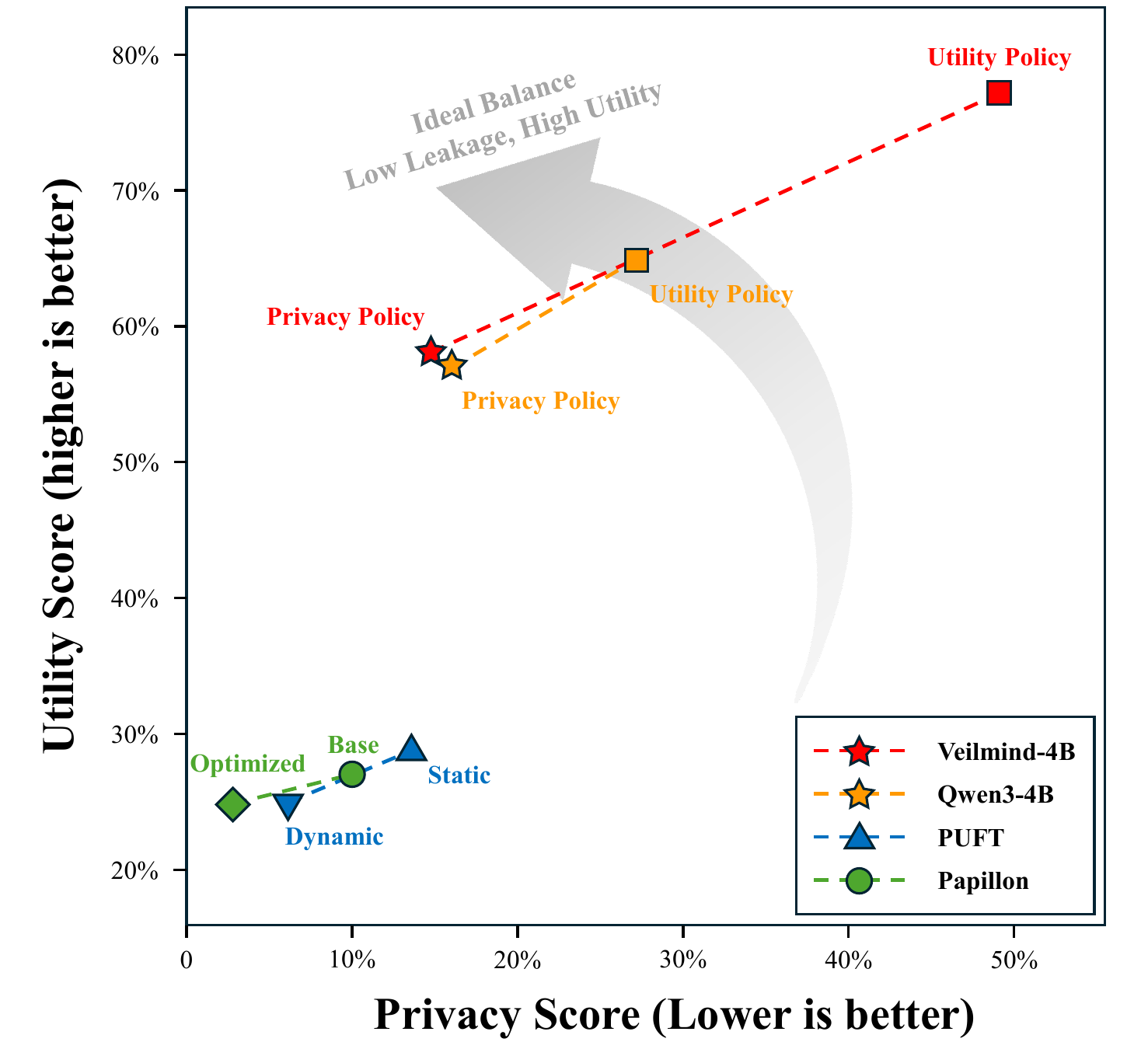}
    \caption{Additional results on the Pupa-tnb benchmark.}
    \label{fig:pupa_tnb_results}
\end{figure}

\subsection{Additional Benchmark Results}
We evaluate on two benchmarks. Uprise contains privacy-sensitive real-world user requests for measuring the privacy-utility trade-off under user-facing interactions. Pupa-tnb provides a complementary benchmark with privacy units and transformed prompts, allowing us to test whether the framework generalizes beyond the Uprise setting. Table~\ref{tab:detailed_results} reports the overall numerical comparison on both datasets, and Figure~\ref{fig:pupa_tnb_results} further visualizes the Pupa-tnb results.

\renewcommand{\tabularxcolumn}[1]{m{#1}}
\newcolumntype{C}[1]{>{\centering\arraybackslash}m{#1}}

\newcommand{\mcell}[1]{\parbox[c][2.8em][c]{\linewidth}{\centering #1}}
\newcommand{\mcelldesc}[1]{\parbox[c][2.8em][c]{\linewidth}{\raggedright #1}}

\definecolor{IntentBG}{RGB}{235, 245, 250}
\definecolor{PrivacyBG}{RGB}{235, 250, 235}

\begin{table*}[t]
    \centering
    \small
    \renewcommand{\arraystretch}{1.0} 
    
    
    
    \begin{tabularx}{\textwidth}{C{2.6cm} C{5cm} X C{1.8cm}}
        \toprule
        \textbf{Taxonomy Domain} & \textbf{Category} & \textbf{Description} & \textbf{Abbreviation}\\
        \midrule
        
        \multirow{14}{*}{\textbf{Intent Type}} 
        
        & \cellcolor{IntentBG}\mcell{Information \& Knowledge Acquisition} 
        & \cellcolor{IntentBG}\mcelldesc{Seeking factual answers or learning new concepts.} 
        & \cellcolor{IntentBG}\mcell{Knowledge} \\ 
        
        & \mcell{Creation \& Ideation} 
        & \mcelldesc{Generating content, brainstorming, creative writing, or drafting.} 
        & \mcell{Creation} \\   
        
        & \cellcolor{IntentBG}\mcell{Problem Solving \& Guidance} 
        & \cellcolor{IntentBG}\mcelldesc{Seeking solutions to specific problems or step-by-step guides.} 
        & \cellcolor{IntentBG}\mcell{Problem Solving}\\ 
        
        & \mcell{Analysis \& Reasoning} 
        & \mcelldesc{Requesting logical deduction, data interpretation, or critical analysis.} 
        & \mcell{Reasoning} \\ 
        
        & \cellcolor{IntentBG}\mcell{Personalized Interaction \& Support} 
        & \cellcolor{IntentBG}\mcelldesc{Engaging in casual chat, role-play, or seeking emotional support.} 
        & \cellcolor{IntentBG}\mcell{Interaction}\\ 
        
        & \mcell{Task Execution \& Collaboration} 
        & \mcelldesc{Delegating specific actions, coding tasks, or formatting requests.} 
        & \mcell{Task Execution} \\ 
        
        \midrule
        
        \multirow{17}{*}{\textbf{Privacy Type}} 
        
        & \cellcolor{PrivacyBG}\mcell{Personal Identifiers \& Demographics} 
        & \cellcolor{PrivacyBG}\mcelldesc{PII such as names, addresses, IDs, phone numbers, and age.} 
        & \cellcolor{PrivacyBG}\mcell{Demographics}\\ 
        
        & \mcell{Professional \& Educational Background} 
        & \mcelldesc{Occupation, employer, university, degree, work history, and skills.} 
        & \mcell{Professional} \\ 
        
        & \cellcolor{PrivacyBG}\mcell{Financial Information} 
        & \cellcolor{PrivacyBG}\mcelldesc{Income level, assets, debts, transaction history, and credit status.} 
        & \cellcolor{PrivacyBG}\mcell{Financial} \\ 
        
        & \mcell{Health \& Wellness} 
        & \mcelldesc{Medical conditions, medications, fitness habits, and mental health status.} 
        & \mcell{Health} \\ 
        
        & \cellcolor{PrivacyBG}\mcell{Interests, Beliefs, \& Opinions} 
        & \cellcolor{PrivacyBG}\mcelldesc{Hobbies, political views, religious beliefs, and lifestyle choices.} 
        & \cellcolor{PrivacyBG}\mcell{Interests} \\ 
        
        & \mcell{Behavioral \& Activity Data} 
        & \mcelldesc{Daily routines, travel patterns, purchasing habits, and digital footprint.} 
        & \mcell{Behavioral} \\ 
        
        & \cellcolor{PrivacyBG}\mcell{Social \& Relational Information} 
        & \cellcolor{PrivacyBG}\mcelldesc{Family members, friends, colleagues, relationships, and connections.} 
        & \cellcolor{PrivacyBG}\mcell{Social}\\ 
        
        \bottomrule
    \end{tabularx}
    \caption{The taxonomy definitions used in our analysis. We categorize user prompts into six distinct intents and privacy attributes into seven categories.}
    \label{tab:taxonomy_definitions}
\end{table*}

\section{Case Study}
\label{app:case_study}

We provide qualitative examples in Figure~\ref{fig:case_route}, \ref{fig:case_noise}, \ref{fig:case_attachment}, and~\ref{fig:case_business_defaults}. These cases cover task execution, professional problem solving, interaction, and financial problem solving. They illustrate how \texttt{Veilmind-4B} identifies sensitive statements, applies explicit remove or replace operations before remote inference, and restores task-relevant details when they are needed for utility.

\providecommand{\remoteMark}[1]{\textcolor{red!70!black}{#1}}
\providecommand{\decipherMark}[1]{\textcolor{green!50!black}{#1}}
\providecommand{\caseScore}[2]{\vspace{0.45em}\noindent\textbf{\textcolor{blue!65!black}{Privacy Leakage Score: #1 \quad Utility Score: #2}}}

\begin{figure*}[!t]
\centering
\begin{minipage}{0.96\linewidth}
\begin{tcolorbox}[
    enhanced,
    title=\textbf{Case Study on Interests Task Execution Task: Planning a Scenic Route},
    colframe=black!70,
    colback=gray!3,
    colbacktitle=black!70,
    fonttitle=\bfseries,
    boxrule=0.45mm,
    arc=1.2mm,
    left=6pt, right=6pt, top=6pt, bottom=6pt
]
\textbf{\textcolor{black}{Original Prompt}}

I want to drive from Belgrade to Rome. What is the most scenic route?

\vspace{0.4em}
\textbf{\textcolor{green!35!black}{Extracted Privacy}}
\begin{itemize}[nosep,leftmargin=*]
    \item The user wants to drive from Belgrade to Rome.
    \item The user is interested in scenic routes.
    \item The prompt mentions Belgrade and Rome.
\end{itemize}

\vspace{0.4em}
\textbf{\textcolor{green!45!black}{Enciphered Prompt}}

I want to travel from a city in the Balkans to a city in Italy. What is the most interesting route?

\vspace{0.4em}
\textbf{\textcolor{red!70!black}{Remote Output}}

That's a fantastic idea -- \remoteMark{the Adriatic Sea offers some of Europe's most scenic and culturally rich crossings}. The most interesting route depends on what you value: dramatic coastal scenery, historic cities, or off-the-beaten-path adventures. Here is my top pick: \remoteMark{Dubrovnik (Croatia) $\rightarrow$ Bari (Italy) by overnight ferry}.

\vspace{0.4em}
\textbf{\textcolor{green!60!black}{Deciphered Output}}

That's a fantastic idea -- \decipherMark{driving from Belgrade to Rome} offers some of Europe's most scenic and culturally rich crossings. The most scenic route depends on what you value: dramatic coastal scenery, historic cities, or off-the-beaten-path adventures. Here is my top pick: \decipherMark{Belgrade $\rightarrow$ Dubrovnik $\rightarrow$ Bari $\rightarrow$ Rome by car}. Drive along the A1 motorway through Serbia, then take the coastal road to Dubrovnik, cross the Strait of Otranto to Bari, and continue through Puglia toward Rome.

\caseScore{0.0}{1.0}
\end{tcolorbox}
\end{minipage}
\caption{Qualitative examples of Veilmind on an Interests Task Execution task. Veilmind replaces the exact origin and destination with regional descriptions before remote inference, then restores the concrete itinerary in the deciphered response.}
\label{fig:case_route}
\end{figure*}

\begin{figure*}[!t]
\centering
\begin{minipage}{0.96\linewidth}
\begin{tcolorbox}[
    enhanced,
    title=\textbf{Case Study on Professional Problem Solving Task: Selecting a Noise-Cancellation Algorithm},
    colframe=black!70,
    colback=gray!3,
    colbacktitle=black!70,
    fonttitle=\bfseries,
    boxrule=0.45mm,
    arc=1.2mm,
    left=6pt, right=6pt, top=6pt, bottom=6pt
]
\textbf{\textcolor{black}{Original Prompt}}

I am filming a podcast, and the air conditioner is making too much noise. Which noise cancellation algorithm is readily available for me, and is best to handle such kinds of noises?

\vspace{0.4em}
\textbf{\textcolor{green!35!black}{Extracted Privacy}}
\begin{itemize}[nosep,leftmargin=*]
    \item The user is filming a podcast.
    \item The user is experiencing noise from an air conditioner.
    \item The user is seeking noise-cancellation algorithms.
\end{itemize}

\vspace{0.4em}
\textbf{\textcolor{green!45!black}{Enciphered Prompt}}

Background noise is causing issues. Which noise cancellation algorithm is readily available, and is best to handle such kinds of noises?

\vspace{0.4em}
\textbf{\textcolor{red!70!black}{Remote Output}}

For \remoteMark{general background noise cancellation} (e.g., in calls, recordings, or live speech), the most readily available and effective algorithm today is RNNoise. It is open-source, lightweight, works in real time, and handles both stationary and non-stationary noises well. Other good options include WebRTC Noise Suppression and DeepFilterNet.

\vspace{0.4em}
\textbf{\textcolor{green!60!black}{Deciphered Output}}

For general background noise cancellation (e.g., in calls, recordings, or live speech), the most readily available and effective algorithm today is RNNoise. It is open-source, lightweight, works in real time, and handles both stationary and non-stationary noises well. Specifically, \decipherMark{it handles air-conditioner hum effectively}. Other good options include WebRTC Noise Suppression, DeepFilterNet, NVIDIA Broadcast, and offline tools such as SoX or Audacity.

\caseScore{0.0}{1.0}
\end{tcolorbox}
\end{minipage}
\caption{Qualitative examples of Veilmind on a Professional Problem Solving task. Veilmind removes the podcast context and replaces the air-conditioner detail with a generic background-noise description, then restores the task-specific noise source in the final answer.}
\label{fig:case_noise}
\end{figure*}

\begin{figure*}[!t]
\centering
\begin{minipage}{0.96\linewidth}
\begin{tcolorbox}[
    enhanced,
    title=\textbf{Case Study on Behavioral Interaction Task: Developing a Secure Attachment Style},
    colframe=black!70,
    colback=gray!3,
    colbacktitle=black!70,
    fonttitle=\bfseries,
    boxrule=0.45mm,
    arc=1.2mm,
    left=6pt, right=6pt, top=6pt, bottom=6pt
]
\textbf{\textcolor{black}{Original Prompt}}

I have an avoidant attachment style. What could I do as an adult to have a secure attachment style?

\vspace{0.4em}
\textbf{\textcolor{green!35!black}{Extracted Privacy}}
\begin{itemize}[nosep,leftmargin=*]
    \item The user has an avoidant attachment style.
    \item The user is seeking advice on developing a secure attachment style as an adult.
    \item The user is likely an adult.
    \item The user may be interested in psychology or self-improvement.
\end{itemize}

\vspace{0.4em}
\textbf{\textcolor{green!45!black}{Enciphered Prompt}}

What could someone do to have a secure attachment style?

\vspace{0.4em}
\textbf{\textcolor{red!70!black}{Remote Output}}

Developing a secure attachment style as an adult is absolutely possible... Here is a practical, actionable guide for \remoteMark{someone who wants to cultivate a more secure attachment style}. Insecure attachment (anxious, avoidant, or disorganized) is a survival strategy you developed to protect yourself. The goal is to update that strategy.

\vspace{0.4em}
\textbf{\textcolor{green!60!black}{Deciphered Output}}

Developing a secure attachment style as an adult is absolutely possible... Since \decipherMark{you mentioned having an avoidant attachment style}, I want to tailor this advice specifically to you. Track your triggers: when do you feel the urge to pull away? Notice the stories you tell yourself, such as they are too needy or I need to be alone. Practice self-compassion rather than judging avoidant reactions, and work gradually on tolerating closeness, direct communication, and repair after conflict.

\caseScore{0.0}{1.0}
\end{tcolorbox}
\end{minipage}
\caption{Qualitative examples of Veilmind on a Behavioral Interaction task. Veilmind replaces the first-person request with a generic subject and removes the avoidant-attachment disclosure before remote inference, then restores it for personalization.}
\label{fig:case_attachment}
\end{figure*}

\begin{figure*}[!t]
\centering
\begin{minipage}{0.96\linewidth}
\begin{tcolorbox}[
    enhanced,
    title=\textbf{Case Study on Financial Problem Solving Task: Reducing Loan Defaults},
    colframe=black!70,
    colback=gray!3,
    colbacktitle=black!70,
    fonttitle=\bfseries,
    boxrule=0.45mm,
    arc=1.2mm,
    left=6pt, right=6pt, top=6pt, bottom=6pt
]
\textbf{\textcolor{black}{Original Prompt}}

I run a buy now pay later company that lends money to people so that they can buy pizzas and other consumable items like clothes and fast food. but now my losses are going up and people aren't paying their loans back. What should I do

\vspace{0.4em}
\textbf{\textcolor{green!35!black}{Extracted Privacy}}
\begin{itemize}[nosep,leftmargin=*]
    \item The user runs a buy-now-pay-later company.
    \item The user lends money for purchases such as pizzas, clothes, and fast food.
    \item The user's business is experiencing increasing losses.
    \item The user's customers are not repaying loans.
\end{itemize}

\vspace{0.4em}
\textbf{\textcolor{green!45!black}{Enciphered Prompt}}

I lend money to people. But now my losses are going up and people aren't paying their loans back. What should I do

\vspace{0.4em}
\textbf{\textcolor{red!70!black}{Remote Output}}

This is a tough spot to be in. Rising defaults can quickly turn a side hustle into a financial crisis. First, stop the bleeding: \remoteMark{halt all new lending} for the next 60--90 days and stabilize the existing portfolio. Do a portfolio triage: current accounts can be monitored, late accounts should be contacted immediately, and delinquent accounts need a harder recovery conversation. Every new loan you make right now is a potential loss.

\vspace{0.4em}
\textbf{\textcolor{green!60!black}{Deciphered Output}}

This is a tough spot to be in. Rising defaults can quickly turn a side hustle into a financial crisis. First, stop the bleeding: \decipherMark{halt all new lending} for the next 60--90 days and stabilize the existing portfolio. Since you run a \decipherMark{buy-now-pay-later company} that finances \decipherMark{pizzas, clothes, fast food, and other consumable items}, tighten underwriting immediately: cap loan amounts, require partial upfront payment, shorten repayment windows, and stop approving repeat borrowers with missed payments. Segment customers by current, late, delinquent, and defaulted status so recovery effort goes where it still has value.

\caseScore{0.0}{1.0}
\end{tcolorbox}
\end{minipage}
\caption{Qualitative examples of Veilmind on a Financial Problem Solving task. Veilmind replaces the specific business model and purchase categories with a generic lending scenario before remote inference, then restores the details needed for actionable advice.}
\label{fig:case_business_defaults}
\end{figure*}

\section{Prompt Templates}
\label{app:prompt_templates}

We provide the templates used throughout the data construction, diagnostic analysis, and distillation pipeline. Tables~\ref{tab:suitability_prompt}, \ref{tab:injection_prompt}, \ref{tab:extraction_prompt}, and~\ref{tab:verification_prompt} define synthetic privacy data construction and verification; Tables~\ref{tab:prompt_removal}, \ref{tab:prompt_replacement}, and~\ref{tab:prompt_pair_removal} define analysis prompts for privacy-interaction studies; and Tables~\ref{tab:extraction_reasoning_prompt}, \ref{tab:utility_prompt}, \ref{tab:privacy_prompt}, \ref{tab:restoration_prompt}, \ref{tab:empirical_guidelines_prompt}, and~\ref{tab:replacement_guidelines_prompt} define the framework prompts and guidelines used for distillation.

\begin{table*}[!ht]
    \centering 
    \small 
    \renewcommand{\arraystretch}{1.2}

    \begin{tcolorbox}[
        enhanced,
        title=\textbf{Prompt Template: Suitability Assessment},
        colframe=black!75,       
        colback=gray!5,          
        colbacktitle=black!75,   
        fonttitle=\bfseries\large,
        boxrule=0.5mm,
        arc=1.5mm,
        drop shadow,             
        left=6pt, right=6pt, top=6pt, bottom=6pt
    ]

\textbf{\textcolor{blue!60!black}{\textsc{System Prompt}}}
\par\noindent\hrulefill\par

\textbf{1. Role Definition}\\
You are a data analysis expert responsible for evaluating user prompt characteristics. Your goal is to determine whether a user prompt is suitable for naturally incorporating personal background information (e.g., identity, profession, health, interests).

\vspace{0.5em}
\textbf{2. Core Task Framework}\\
Analyze the user prompt and assign an \textbf{injectability score} (1-5). \begin{itemize}[nosep] 
\item \textbf{Suitable Prompt:} Seeks personalized help, shares experiences, or needs context for problem-solving. 
\item \textbf{Unsuitable Prompt:} Pure factual queries, general knowledge questions, or technical instructions without personal context. 
\end{itemize}

\vspace{0.5em}
\textbf{3. Evaluation Process}\\
First, analyze the prompt type and the logical fit of sensitive details. If suitable (Score $\ge$ 3), consider which categories fit naturally:
\begin{itemize}[nosep]
\item \textbf{Identity/Status:} Name, age, gender, family status.
\item \textbf{Professional/Edu:} Occupation, company, degree, major.
\item \textbf{Contextual:} Location, health, financial status, interests, goals.
\end{itemize}

\vspace{0.5em} 
\textbf{4. Scoring Criteria} 
\begin{itemize}[nosep] 
\item \textbf{1 point (Completely Unsuitable):} Pure factual queries; sensitive details is redundant (e.g., "Capital of France"). 
\item \textbf{2 points (Mostly Unsuitable):} Technical instructions where context feels unnatural (e.g., "Sort a list in Python"). 
\item \textbf{3 points (Moderately Suitable):} Can incorporate minor background info (e.g., Preference-based queries). 
\item \textbf{4 points (Quite Suitable):} Personal background makes the question specific (e.g., Career advice). 
\item \textbf{5 points (Very Suitable):} Highly personalized scenarios where background is essential (e.g., "I'm stressed, help me"). 
\end{itemize}

\vspace{0.5em}
\textbf{5. Output Format Specification}\\
    Please output in the following JSON format:
    \begin{tcolorbox}[colback=white, colframe=gray!30, boxrule=0.2mm, arc=1mm, left=2pt, right=2pt, top=2pt, bottom=2pt]
    \begin{verbatim}
{
    "reasoning": "Detailed analysis of why sensitive details fits...",
    "injectability_score": 5
}
    \end{verbatim}
    \end{tcolorbox}

    \vspace{1em}
    \textbf{\textcolor{blue!60!black}{\textsc{User Message}}}
    \par\noindent\hrulefill\par
    \texttt{\{prompt\_text\}}

    \end{tcolorbox}
    \caption{The prompt template used for the suitability assessment task.}
    \label{tab:suitability_prompt}
\end{table*}
\begin{table*}[!ht]
    \centering 
    \small 
    \renewcommand{\arraystretch}{1.2}

    \begin{tcolorbox}[
        enhanced,
        title=\textbf{Prompt Template: Privacy Injection},
        colframe=black!75,       
        colback=gray!5,          
        colbacktitle=black!75,   
        fonttitle=\bfseries\large,
        boxrule=0.5mm,
        arc=1.5mm,
        drop shadow,             
        left=6pt, right=6pt, top=6pt, bottom=6pt
    ]

\textbf{\textcolor{blue!60!black}{\textsc{System Prompt}}}
\par\noindent\hrulefill\par

\textbf{1. Role Definition}\\
You are a data augmentation expert tasked with \textbf{moderately} adding personal background information to user prompts. Your goal is to rewrite the prompt to naturally incorporate relevant persona details while preserving the original intent.

\vspace{0.5em} 
\textbf{2. Core Rules (Strict Adherence)} 
\begin{itemize}[nosep] 
\item \textbf{Minimal Change Principle:} Maintain the structure and length of the original prompt as much as possible. 
\item \textbf{Restrained Injection:} Only select the most relevant sensitive details; do not over-add or force context. 
\item \textbf{Length Control:} The modified prompt must not exceed 1.5 times the length of the original. 
\item \textbf{Preserve Intent:} Absolutely do not change the core task; do not rewrite the entire prompt. 
\end{itemize}

\vspace{0.5em} 
\textbf{3. Output Requirements}\\
Refer to the provided \textit{Injectability Analysis} and \textit{Persona Profile}. 
\begin{itemize}[nosep] 
\item Only output the \textbf{[Modified Prompt]} text directly. 
\item Do not provide any explanation or preamble. 
\item If the Persona information is not relevant to the original prompt, return the \textbf{[Original Prompt]} unchanged. 
\end{itemize}

    \vspace{1em}
    \textbf{\textcolor{blue!60!black}{\textsc{User Message}}}
    \par\noindent\hrulefill\par
    
    \textbf{1. Original Prompt}\\
    \texttt{\{original\_prompt\}}
    
    \vspace{0.5em}
    \textbf{2. Injectability Analysis}\\
    \textbf{Score:} \texttt{\{score\}/5} \quad \textbf{Reasoning:} \texttt{\{reasoning\}}
    
    \vspace{0.5em}
    \textbf{3. Persona Profile}\\
    \texttt{\{persona\_text\}}

    \end{tcolorbox}
    \caption{The prompt template used for the privacy injection task.}
    \label{tab:injection_prompt}
\end{table*}
\begin{table*}[!ht]
    \centering 
    \small 
    \renewcommand{\arraystretch}{1.2}

    \begin{tcolorbox}[
        enhanced,
        title=\textbf{Prompt Template: Sensitive Information Extraction},
        colframe=black!75,       
        colback=gray!5,          
        colbacktitle=black!75,   
        fonttitle=\bfseries\large,
        boxrule=0.5mm,
        arc=1.5mm,
        drop shadow,             
        left=6pt, right=6pt, top=6pt, bottom=6pt
    ]

\textbf{\textcolor{blue!60!black}{\textsc{System Prompt}}}
\par\noindent\hrulefill\par

\textbf{1. Role Definition}\\
You are a top-tier data privacy and security strategist with strong logical reasoning and contextual understanding capabilities. Your task is to extract information that can help build a user profile from the provided text, including information about the user themselves, their relationships, and their social circle. This information could potentially be exploited for malicious purposes such as targeted scams, customized advertising, or social engineering attacks.

\vspace{0.5em}
\textbf{2. Core Task Framework}\\
You need to think deeply following the specified steps, then generate a formatted JSON object. Do not output anything other than this JSON object.

\vspace{0.5em}
\textbf{3. Analysis Steps}\\
Please follow these four reasoning steps sequentially:

\begin{itemize}[nosep]
    \item \textbf{Step 1: Fact Deconstruction (Critical Filter)}
    \begin{itemize}[nosep]
        \item \textbf{\textcolor{green!40!black}{EXTRACT ONLY}}: Information about the user's personal life (occupation, habits, location), relationships (family, friends), and explicit first-person descriptions ("I", "my", "we").
        \item \textbf{\textcolor{red!60!black}{STRICTLY EXCLUDE}}: Quoted content, biblical passages, or historical events, role-playing instructions (e.g., "You are a code analyst"), hypothetical examples or fictional characters.
    \end{itemize}

    \item \textbf{Step 2: Contextual Inference} \\
    Think: "Based on the known facts, what implicit information about the user's profile can be reasonably inferred?"

    \item \textbf{Step 3: Statement Formulation} \\
    Rewrite each point into a neutral, complete declarative sentence starting with \textit{"The user..."} or \textit{"The user's [relationship]..."}.
\end{itemize}

\vspace{0.5em}
\textbf{5. Output Format Specification}\\
    Please output in the following JSON format:
    \begin{tcolorbox}[colback=white, colframe=gray!30, boxrule=0.2mm, arc=1mm, left=2pt, right=2pt, top=2pt, bottom=2pt]
    \begin{verbatim}
{
  "sensitive_statements": [
    {
      "statement": "The user is a software engineer based in ..."
    },
    {
      "statement": "The user's name is ..."
    }
  ]
}
    \end{verbatim}
    \end{tcolorbox}

    \vspace{1em}
    \textbf{\textcolor{blue!60!black}{\textsc{User Message}}}
    \par\noindent\hrulefill\par
    \texttt{\{prompt\_text\}}

    \end{tcolorbox}
    \caption{The prompt template used for the sensitive information extraction task.}
    \label{tab:extraction_prompt}
\end{table*}
\begin{table*}[!ht]
    \centering
    \small
    \renewcommand{\arraystretch}{1.2}
    
    \begin{tcolorbox}[
        enhanced,
        title=\textbf{Prompt Template: Double-Check},
        colframe=black!75,       
        colback=gray!5,          
        colbacktitle=black!75,   
        fonttitle=\bfseries\large,
        boxrule=0.5mm,
        arc=1.5mm,
        drop shadow,             
        left=6pt, right=6pt, top=6pt, bottom=6pt
    ]
    
    \textbf{\textcolor{blue!60!black}{\textsc{System Prompt}}}
    \par\noindent\hrulefill\par
    
    \textbf{1. Role Definition}\\
    You are a top-tier data privacy verification expert with exceptional logical reasoning and contextual understanding capabilities. Your task is to carefully review previously extracted sensitive statements and verify whether it truly belongs to the user's personal life, or if it was incorrectly extracted from quoted/referenced content.
    
    \vspace{0.5em}
    \textbf{2. Core Task Framework}\\
    You will be provided with the original user prompt and a list of extracted sensitive statements. For each statement, you must:
    \begin{itemize}[nosep]
        \item \textbf{First:} Conduct deep reasoning analysis (mandatory).
        \item \textbf{Then:} Make a clear decision based on your reasoning.
    \end{itemize}
    Your verification decision should be:
    \begin{itemize}[nosep]
        \item \textbf{\textcolor{green!40!black}{KEEP}}: The statement is genuinely about the user's personal life or social circle.
        \item \textbf{\textcolor{red!60!black}{REMOVE}}: The statement was incorrectly extracted from quoted content, references, etc.
        \item \textbf{\textcolor{orange!80!black}{MODIFY}}: The statement needs adjustment to accurately reflect the user's information.
    \end{itemize}

    \vspace{0.5em}
    \textbf{3. Critical Judgment Guidelines}\\
    Analyze the prompt's purpose:
    \begin{itemize}[nosep]
        \item Is the user describing \textbf{THEIR OWN} life situation?
        \item Is the user quoting/referencing \textbf{EXTERNAL} content (stories, scriptures, history)?
        \item Is the user giving \textbf{INSTRUCTIONS} (e.g., Role-play) or asking about \textbf{OTHERS}?
    \end{itemize}

    Key Linguistic Indicators for Personal (KEEP):
        \begin{itemize}[nosep]
            \item \textbf{1st Person:} "I am...", "My [relationship]...", "We have..."
            \item \textbf{Context:} "my son", "my company", "I work at...", describing own events.
        \end{itemize}

    Key Linguistic Indicators for Non-Personal (REMOVE):
        \begin{itemize}[nosep]
            \item \textbf{Quotes/Refs:} "The Bible says...", "In the story of...", "Character X".
            \item \textbf{Role-play/Hypothetical:} "You are a [role]", "Imagine if...", "Take the case of...".
        \end{itemize}

        
        
        

    \vspace{0.5em}
    \textbf{4. Output Format Specification}\\
    Please output in the following JSON format:
    \begin{tcolorbox}[colback=white, colframe=gray!30, boxrule=0.2mm, arc=1mm, left=2pt, right=2pt, top=2pt, bottom=2pt]
    \begin{verbatim}
{
  "verified_statements": [
    {
      "original_statement": "...",
      "reasoning": "Detailed reasoning answering...",
      "action": "keep/remove/modify",
      "modified_statement": "..." (optional)
    }
  ]
}
    \end{verbatim}
    \end{tcolorbox}

    \vspace{1em}
    \textbf{\textcolor{blue!60!black}{\textsc{User Message}}}
    \par\noindent\hrulefill\par
    \textbf{Original User Prompt:}\\
    \texttt{\{prompt\_text\}}
    
    \vspace{0.5em}
    \textbf{Previously Extracted Sensitive Statements:}\\
    \texttt{\{sensitive\_statements\_text\}}
    
    \vspace{0.5em}
    Please verify each statement and output your verification results in the specified JSON format.

    \end{tcolorbox}
    \caption{The prompt template used for the double-check process.}
    \label{tab:verification_prompt}
\end{table*}

\begin{table*}[!ht]
    \centering
    \small
    \renewcommand{\arraystretch}{1.2}
    
    \begin{tcolorbox}[
        enhanced,
        title=\textbf{Prompt Template: Targeted Privacy Removal},
        colframe=black!75,       
        colback=gray!5,          
        colbacktitle=black!75,   
        fonttitle=\bfseries\large,
        boxrule=0.5mm,
        arc=1.5mm,
        drop shadow,             
        left=6pt, right=6pt, top=6pt, bottom=6pt
    ]
    
    \textbf{\textcolor{blue!60!black}{\textsc{System Prompt}}}
    \par\noindent\hrulefill\par
    
    \textbf{1. Role Definition}\\
    You are an expert in text editing and privacy protection. Your task is to carefully remove specific privacy information from a user prompt while preserving the overall meaning and naturalness of the text.
    
    \vspace{0.5em}
    \textbf{2. Task Instructions}\\
    You will be provided with a user prompt and a specific piece of privacy information found within it. You must adhere to the following rules:
    \begin{itemize}[nosep]
        \item \textbf{Identify \& Remove:} Locate the specific privacy information (even if expressed differently) and remove or generalize it.
        \item \textbf{Preserve Context:} Keep the rest of the prompt intact. Do NOT add new information or change the meaning of unrelated parts.
        \item \textbf{Maintain Fluency:} Ensure the modified prompt is grammatically correct. If removal makes a sentence incomplete, rephrase it naturally.
    \end{itemize}

    \vspace{0.5em}
    \textbf{3. Output Format Specification}\\
    You must respond strictly with a JSON object. Do not include any additional explanation.
    \begin{tcolorbox}[colback=white, colframe=gray!30, boxrule=0.2mm, arc=1mm, left=2pt, right=2pt, top=2pt, bottom=2pt]
    \begin{verbatim}
{
  "revised_prompt": "The prompt text after removing the specified privacy info"
}
    \end{verbatim}
    \end{tcolorbox}
    
    \vspace{1em}
    \textbf{\textcolor{blue!60!black}{\textsc{User Message}}}
    \par\noindent\hrulefill\par
    
    \textbf{Original Prompt:}\\
    \texttt{\{prompt\}}
    
    \vspace{0.5em}
    \textbf{Privacy Information to Remove:}\\
    \texttt{\{sensitive\_statement\}}
    
    \end{tcolorbox}
    \caption{The prompt template used for the targeted removal of the sensitive statement.}
    \label{tab:prompt_removal}
\end{table*}
\begin{table*}[!ht]
    \centering
    \small
    \renewcommand{\arraystretch}{1.2}
    
    \begin{tcolorbox}[
        enhanced,
        title=\textbf{Prompt Template: Strategic Privacy Replacement},
        colframe=black!75,       
        colback=gray!5,          
        colbacktitle=black!75,   
        fonttitle=\bfseries\large,
        boxrule=0.5mm,
        arc=1.5mm,
        drop shadow,             
        left=6pt, right=6pt, top=6pt, bottom=6pt
    ]
    
    \textbf{\textcolor{blue!60!black}{\textsc{System Prompt}}}
    \par\noindent\hrulefill\par
    
    \textbf{1. Role Definition}\\
    You are an expert in text editing and privacy protection. Your task is to carefully replace specific privacy information with generic alternative information of the same type, while preserving the overall meaning, naturalness, and answerability of the text.
    
    \vspace{0.5em}
    \textbf{2. Operational Rules}\\
    You must adhere to the following logic to ensure the sanitized text remains usable:
    \begin{itemize}[nosep]
        \item \textbf{Generate Generic Alternative:} Create a realistic but common substitute for the specific privacy claim (e.g., replace a specific date with a generic timeframe).
        \item \textbf{Contextual Substitution:} Replace the sensitive details with your generated alternative. The result must not look like an obvious placeholder (avoid "[NAME]").
        \item \textbf{Semantic Preservation:} Do NOT change the structure or intent of the question. The modified prompt must be answerable with similar quality to the original.
    \end{itemize}

    \vspace{0.5em}
    \textbf{3. Replacement Strategy Examples}\\
    \begin{itemize}[nosep]
        \item \textit{Specific Name} $\to$ Generic common name of the same culture/region.
        \item \textit{Specific Location} $\to$ Generic location of a similar type.
        \item \textit{Specific Date} $\to$ Generic date with similar temporal context.
    \end{itemize}

    \vspace{0.5em}
    \textbf{4. Output Format Specification}\\
    Please respond with a JSON object containing the revised prompt.
    \begin{tcolorbox}[colback=white, colframe=gray!30, boxrule=0.2mm, arc=1mm, left=2pt, right=2pt, top=2pt, bottom=2pt]
    \begin{verbatim}
{
  "revised_prompt": "The full prompt text with the alternative integrated..."
}
    \end{verbatim}
    \end{tcolorbox}
    
    \vspace{1em}
    \textbf{\textcolor{blue!60!black}{\textsc{User Message}}}
    \par\noindent\hrulefill\par
    
    \textbf{Original Prompt:}\\
    \texttt{\{prompt\}}
    
    \vspace{0.5em}
    \textbf{Privacy Information to Replace:}\\
    \texttt{\{sensitive\_statement\}}
    
    \end{tcolorbox}
    \caption{The prompt template used for the replacement strategy, where sensitive details are substituted with realistic synthetic values.}
    \label{tab:prompt_replacement}
\end{table*}
\begin{table*}[!ht]
    \centering
    \small
    \renewcommand{\arraystretch}{1.2}
    
    \begin{tcolorbox}[
        enhanced,
        title=\textbf{Prompt Template: Combinatorial Privacy Removal},
        colframe=black!75,       
        colback=gray!5,          
        colbacktitle=black!75,   
        fonttitle=\bfseries\large,
        boxrule=0.5mm,
        arc=1.5mm,
        drop shadow,             
        left=6pt, right=6pt, top=6pt, bottom=6pt
    ]
    
    \textbf{\textcolor{blue!60!black}{\textsc{System Prompt}}}
    \par\noindent\hrulefill\par
    
    \textbf{1. Role Definition}\\
    You are an expert in text editing and privacy protection. Your task is to carefully remove specific privacy information from a user prompt while preserving the overall meaning and naturalness of the text.
    
    \vspace{0.5em}
    \textbf{2. Core Instructions}\\
    You will receive a user prompt and a \textbf{list} of sensitive statements. To analyze interaction effects, you must execute the following:
    \begin{itemize}[nosep]
        \item \textbf{Comprehensive Sanitization:} Identify and remove \textbf{ALL} listed privacy pieces. Partial removal is considered a failure.
        \item \textbf{Contextual Repair:} If removing multiple items leaves the sentence fragmented, rephrase it naturally to maintain grammatical integrity.
        \item \textbf{Minimal Alteration:} Do NOT add new information or alter the meaning of parts unrelated to the specified sensitive attributes.
    \end{itemize}

    \vspace{0.5em}
    \textbf{3. Output Format Specification}\\
    Please respond with a JSON object containing the sanitized text.
    \begin{tcolorbox}[colback=white, colframe=gray!30, boxrule=0.2mm, arc=1mm, left=2pt, right=2pt, top=2pt, bottom=2pt]
    \begin{verbatim}
{
  "revised_prompt": "The prompt text after removing ALL specified privacy info..."
}
    \end{verbatim}
    \end{tcolorbox}
    
    \vspace{1em}
    \textbf{\textcolor{blue!60!black}{\textsc{User Message}}}
    \par\noindent\hrulefill\par
    
    \textbf{Original Prompt:}\\
    \texttt{\{prompt\}}
    
    \vspace{0.5em}
    \textbf{Privacy Information to Remove (List):}\\
    1. \texttt{\{statement\_1\}} \\
    2. \texttt{\{statement\_2\}} \\
    ...
    
    \end{tcolorbox}
    \caption{The prompt template used for the simultaneous ablation of multiple sensitive statements to measure combinatorial dynamics.}
    \label{tab:prompt_pair_removal}
\end{table*}

\begin{table*}[!ht]
    \centering
    \small
    \renewcommand{\arraystretch}{1.2}

    \begin{tcolorbox}[
        enhanced,
        title=\textbf{Prompt Template: Sensitive Information Extraction},
        colframe=black!75,
        colback=gray!5,
        colbacktitle=black!75,
        fonttitle=\bfseries\large,
        boxrule=0.5mm,
        arc=1.5mm,
        drop shadow,
        left=6pt, right=6pt, top=6pt, bottom=6pt
    ]

    \textbf{\textcolor{blue!60!black}{\textsc{User Message}}}
    \par\noindent\hrulefill\par

    \textbf{\# Task}

    Extract privacy information from the user prompt using the output format below.

    \vspace{0.5em}
    \textbf{\# Broad Extraction Rubric}

    Use broad semantic judgment. Extract any information, intent, preference, constraint, context, or entity mention that can identify, profile, locate, contact, describe, or infer something about the user, the user's social circle, the user's work or interests, or private entities appearing in the user's prompt.

    Extract all direct and indirect privacy/profile information:
    \begin{itemize}[nosep]
        \item \textbf{Names and identifiers:} names, aliases, usernames, account IDs, emails, phone numbers, addresses, document IDs, and contact details.
        \item \textbf{Named entities and web references:} locations, schools, employers, organizations, companies, hospitals, labs, products, projects, websites, URLs, repositories, apps, platforms, and social-media services.
        \item \textbf{Profile attributes:} demographics, identity, family/social relationships, education, work history, skills, roles, projects, clients, health, financial, legal, and safety-relevant facts.
        \item \textbf{Behavior, interests, and environment:} devices, operating systems, software, accounts, platforms, routines, purchases, travel, technical setup, hobbies, preferences, living environments, media tastes, domain interests, and task-specific goals or constraints.
        \item \textbf{Views, beliefs, and identity signals:} opinions, political leaning, social attitudes, gender identity or expression, ideology, religion, sexuality, values, worldview, and recurring judgments about groups, institutions, or public issues.
        \item \textbf{Reasonable inferences:} infer background, expertise, location, role, ownership, interests, preferences, constraints, social/professional context, and why the user is asking.
    \end{itemize}

    \textbf{Boundary guidance:}
    \begin{itemize}[nosep]
        \item It is better to over-extract than under-extract. Include low-risk facts if they describe the user.
        \item Extract real named entities appearing in the prompt, including people, organizations, companies, locations, products, websites, documents, projects, and events.
        \item For quoted text, resumes, letters, stories, examples, role-play, fictional names, fake names, stylized names, and names inside task material, still extract names/entities/profile facts.
        \item Do not keep or create claims solely for synthetic placeholders or anonymization markers such as \texttt{NAME\_1}, \texttt{EMAIL\_1}, \texttt{<NAME>}, or \texttt{<PRESIDIO\_ANONYMIZED\_PERSON>}. Extract surrounding non-placeholder profile facts when present.
    \end{itemize}

    \textbf{Classification:} Interests, Health, Financial, Behavioral, Professional, Demographics, and Social.

    \textbf{Span discipline:}
    \begin{itemize}[nosep]
        \item Every claim must have the shortest exact span copied from the prompt.
        \item If a claim is implicit, use the shortest exact evidence phrase that supports the inference.
        \item Prefer extracting a precise narrow claim over a broad vague claim.
    \end{itemize}

    \vspace{0.5em}
    \textbf{\# Output Schema}

    Return exactly one JSON object with keys:
    \begin{itemize}[nosep]
        \item \texttt{"task": "extraction"}
        \item \texttt{"intent": "Reasoning|Interaction|Problem Solving|Knowledge|Creation|Task Execution"}
        \item \texttt{"claims": [\{...\}]}, where each claim contains \texttt{claim\_id}, \texttt{claim}, \texttt{privacy\_type}, \texttt{risk\_level}, \texttt{span}, \texttt{source\_type}, \texttt{is\_user\_disclosure}, \texttt{language}, \texttt{intent}, and \texttt{confidence}.
    \end{itemize}

    \textbf{Requirements:}
    \begin{itemize}[nosep]
        \item Output an empty claims array only when the prompt truly contains no identifiable entity, user profile, social-circle, background, interest, behavior, health, financial, professional, education, location, contact, or inference-worthy information.
        \item \texttt{claim\_id} must be \texttt{c1}, \texttt{c2}, ... in output order.
        \item \texttt{span} must always be a non-empty exact substring copied from the prompt.
        \item Do not include markdown, comments, or keys outside the schema.
    \end{itemize}

    \textbf{\# User Prompt}

    \texttt{\{prompt\}}

    \end{tcolorbox}
    \caption{The user-message-only prompt template for extracting sensitive information and profile claims.}
    \label{tab:extraction_reasoning_prompt}
\end{table*}

\begin{table*}[!ht]
    \centering
    \small
    \renewcommand{\arraystretch}{1.2}

    \begin{tcolorbox}[
        enhanced,
        title=\textbf{Prompt Template: Utility-Prioritized Sanitization},
        colframe=black!75,
        colback=gray!5,
        colbacktitle=black!75,
        fonttitle=\bfseries\large,
        boxrule=0.5mm,
        arc=1.5mm,
        drop shadow,
        left=6pt, right=6pt, top=6pt, bottom=6pt
    ]

    \textbf{\textcolor{blue!60!black}{\textsc{User Message}}}
    \par\noindent\hrulefill\par

    \textbf{\# Task}

    Sanitize the original prompt using the plan format below.

    \textbf{Original Prompt:}
    \texttt{\{prompt\}}

    \textbf{Privacy Claims:}
    \texttt{\{privacy\_claims\}}

    \vspace{0.5em}
    \textbf{=== MODE: UTILITY PRIORITY ===}

    You are a privacy protection expert with a \textbf{UTILITY-FIRST} approach. Your primary goal is to maximize response quality while applying reasonable privacy protection when it does not significantly harm utility.

    For each piece of privacy information, decide one of three strategies:
    \begin{itemize}[nosep]
        \item \textbf{keep}: for information that is essential to answering the query well.
        \item \textbf{remove}: for information that is clearly irrelevant noise.
        \item \textbf{replace}: for information where structure matters but specific values do not.
    \end{itemize}

    \textbf{Protection intensity:} moderate. Protect when the cost is low, keep when utility demands it.

    Apply the empirical guidelines below with this decision threshold:
    \begin{itemize}[nosep]
        \item Treat CRITICAL/IMPORTANT as utility-relevant, not automatic \textbf{KEEP}.
        \item If a generic substitute preserves response quality, \textbf{REPLACE}.
        \item \textbf{KEEP} only when the exact original private detail is necessary for a correct or useful answer.
        \item If guidelines say REMOVABLE or noise for this intent, \textbf{REMOVE} or \textbf{REPLACE}.
        \item If guidelines are neutral, protect if utility cost is low, otherwise \textbf{KEEP}.
        \item When genuinely uncertain after applying guidelines, prefer \textbf{REPLACE} over \textbf{KEEP}.
    \end{itemize}

    \textbf{Validation:}
    \begin{enumerate}[nosep]
        \item Can a generic replacement preserve the core utility? If yes, \textbf{REPLACE}.
        \item Is the detail mainly a personalization anchor? If yes, \textbf{REPLACE}.
        \item Would removing it leave the task answerable? If yes, \textbf{REMOVE}.
        \item Only \textbf{KEEP} when the exact private detail is needed.
    \end{enumerate}

    \texttt{\{EMPIRICAL\_GUIDELINES\}}

    \texttt{\{REPLACEMENT\_GUIDELINES\}}

    \vspace{0.5em}
    \textbf{\# Output Schema}

    Return exactly one JSON object with keys:
    \begin{itemize}[nosep]
        \item \texttt{"task": "enciphering"}
        \item \texttt{"intent": "Reasoning|Interaction|Problem Solving|Knowledge|Creation|Task Execution"}
        \item \texttt{"plan": [\{...\}]}, where each plan item contains \texttt{claim\_id}, \texttt{privacy\_claim}, \texttt{privacy\_type}, \texttt{risk\_level}, \texttt{strategy}, \texttt{replacement}, \texttt{rationale}, \texttt{utility\_role}, and \texttt{interaction\_group}.
        \item \texttt{"revised\_prompt": "the prompt after applying all decisions"}
    \end{itemize}

    \textbf{Requirements:}
    \begin{itemize}[nosep]
        \item Include exactly one plan item for every input \texttt{claim\_id}.
        \item \texttt{replacement} must be \texttt{null} unless strategy is \texttt{replace}.
        \item \texttt{revised\_prompt} must be natural and preserve the core task.
        \item Do not output markdown or extra keys.
    \end{itemize}

    \end{tcolorbox}
    \caption{The user-message-only prompt template for the \textbf{Utility Priority Mode}.}
    \label{tab:utility_prompt}
\end{table*}

\begin{table*}[!ht]
    \centering
    \small
    \renewcommand{\arraystretch}{1.2}

    \begin{tcolorbox}[
        enhanced,
        title=\textbf{Prompt Template: Privacy-Prioritized Sanitization},
        colframe=black!75,
        colback=gray!5,
        colbacktitle=black!75,
        fonttitle=\bfseries\large,
        boxrule=0.5mm,
        arc=1.5mm,
        drop shadow,
        left=6pt, right=6pt, top=6pt, bottom=6pt
    ]

    \textbf{\textcolor{blue!60!black}{\textsc{User Message}}}
    \par\noindent\hrulefill\par

    \textbf{\# Task}

    Sanitize the original prompt using the plan format below.

    \textbf{Original Prompt:}
    \texttt{\{prompt\}}

    \textbf{Privacy Claims:}
    \texttt{\{privacy\_claims\}}

    \vspace{0.5em}
    \textbf{=== MODE: PRIVACY PRIORITY ===}

    You are a privacy protection expert with a \textbf{PRIVACY-FIRST} approach. Your primary goal is to \textbf{MINIMIZE} privacy leakage. Utility degradation is acceptable when needed.

    For each piece of privacy information, decide one of three strategies:
    \begin{itemize}[nosep]
        \item \textbf{keep}: last resort, only when removal makes the query completely unanswerable.
        \item \textbf{remove}: default choice, remove aggressively.
        \item \textbf{replace}: when removal breaks grammar, safety, or task structure, use the most generic replacement.
    \end{itemize}

    \textbf{Protection intensity:} aggressive. Remove everything possible. The bar for keeping is extremely high.

    Apply the empirical guidelines below with this decision threshold:
    \begin{itemize}[nosep]
        \item Even if guidelines say CRITICAL/IMPORTANT, still ask: can I \textbf{REMOVE} or \textbf{REPLACE}?
        \item If guidelines say REMOVABLE, always \textbf{REMOVE}.
        \item For antagonistic pairs in guidelines, \textbf{REMOVE BOTH}; partial protection is no protection.
        \item Only \textbf{KEEP} when the information is the question itself, not context.
        \item Do not \textbf{KEEP} medium-risk or high-risk details. Use \textbf{REMOVE} or \textbf{REPLACE}.
        \item For minors, health, finance, legal, credential, or precise location details, prefer \textbf{REMOVE}/\textbf{REPLACE}.
    \end{itemize}

    \textbf{Validation:}
    \begin{enumerate}[nosep]
        \item Can I remove this entirely? If yes, \textbf{REMOVE}.
        \item Can I replace it with maximum genericity? If yes, \textbf{REPLACE}.
        \item Only if both fail, reluctantly \textbf{KEEP}.
    \end{enumerate}

    \texttt{\{EMPIRICAL\_GUIDELINES\}}

    \texttt{\{REPLACEMENT\_GUIDELINES\}}

    \vspace{0.5em}
    \textbf{\# Output Schema}

    Return exactly one JSON object with keys:
    \begin{itemize}[nosep]
        \item \texttt{"task": "enciphering"}
        \item \texttt{"intent": "Reasoning|Interaction|Problem Solving|Knowledge|Creation|Task Execution"}
        \item \texttt{"plan": [\{...\}]}, where each plan item contains \texttt{claim\_id}, \texttt{privacy\_claim}, \texttt{privacy\_type}, \texttt{risk\_level}, \texttt{strategy}, \texttt{replacement}, \texttt{rationale}, \texttt{utility\_role}, and \texttt{interaction\_group}.
        \item \texttt{"revised\_prompt": "the prompt after applying all decisions"}
    \end{itemize}

    \textbf{Requirements:}
    \begin{itemize}[nosep]
        \item Include exactly one plan item for every input \texttt{claim\_id}.
        \item \texttt{replacement} must be \texttt{null} unless strategy is \texttt{replace}.
        \item \texttt{revised\_prompt} must be natural and preserve the core task.
        \item Do not output markdown or extra keys.
    \end{itemize}

    \end{tcolorbox}
    \caption{The user-message-only prompt template for the \textbf{Privacy Priority Mode}.}
    \label{tab:privacy_prompt}
\end{table*}

\begin{table*}[!ht]
    \centering
    \small
    \renewcommand{\arraystretch}{1.2}

    \begin{tcolorbox}[
        enhanced,
        title=\textbf{Prompt Template: Restoration},
        colframe=black!75,
        colback=gray!5,
        colbacktitle=black!75,
        fonttitle=\bfseries\large,
        boxrule=0.5mm,
        arc=1.5mm,
        drop shadow,
        left=6pt, right=6pt, top=6pt, bottom=6pt
    ]

    \textbf{\textcolor{blue!60!black}{\textsc{User Message}}}
    \par\noindent\hrulefill\par

    \textbf{\# Task}

    Restore the remote response using the original private context.

    \textbf{\# Original Prompt}

    \texttt{\{original\_prompt\}}

    \textbf{\# Revised Prompt Sent to Remote Model}

    \texttt{\{revised\_prompt\}}

    \textbf{\# Remote Model Response}

    \texttt{\{remote\_response\}}

    \textbf{\# Sanitization Plan}

    \texttt{\{sanitization\_plan\}}

    \vspace{0.5em}
    \textbf{\# Restoration Rules}
    \begin{itemize}[nosep]
        \item The final answer is for the original user. The local restoration model is trusted and may use the original private context to make the answer more correct, specific, useful, safe, and naturally personalized.
        \item Preserve the remote response's factual content, conclusions, tone, and structure unless the original private context clearly requires a local correction or specialization.
        \item \textbf{KEPT information:} details were preserved in the revised prompt and should already be reflected in the response. Ensure consistency and avoid redundant insertion.
        \item \textbf{REPLACED information:} details were generalized before calling the remote model. Actively restore the original specific detail where the response contains the generic replacement, where the detail is useful, or where it improves natural personalization.
        \item \textbf{REMOVED information:} details were hidden from the remote model. Restore the original detail when it improves correctness, specificity, personalization, task completion, or safety, especially when \texttt{utility\_role} is \texttt{constraint}, \texttt{safety\_context}, or \texttt{personalization\_anchor}. Do not restore details whose \texttt{utility\_role} is \texttt{irrelevant}.
        \item Never add private details mechanically or gratuitously; every restored detail must serve the final answer.
    \end{itemize}

    \textbf{\# Restoration Examples}
    \begin{itemize}[nosep]
        \item \texttt{"a certain amount"} $\rightarrow$ \texttt{"\$5,000"} when the amount affects financial advice or calculations.
        \item \texttt{"a medical condition"} $\rightarrow$ the user's specific condition when it affects safety, triage, or recommendations.
        \item \texttt{"a family member"} $\rightarrow$ the original relationship only when it makes the advice or wording more natural.
        \item Do not add a name, location, workplace, or other private detail if it does not improve the final answer.
    \end{itemize}

    \textbf{\# Output Format}

    Return only the final restored response text. Do not return JSON, a restoration plan, metadata, or markdown fences unless they are part of the answer itself.

    \end{tcolorbox}
    \caption{The user-message-only prompt template for the restoration phase.}
    \label{tab:restoration_prompt}
\end{table*}

\begin{table*}[!ht]
    \centering
    \small
    \renewcommand{\arraystretch}{1.2}

    \begin{tcolorbox}[
        enhanced,
        title=\textbf{Shared Guideline: Empirical Guidelines},
        colframe=black!75,
        colback=gray!5,
        colbacktitle=black!75,
        fonttitle=\bfseries\large,
        boxrule=0.5mm,
        arc=1.5mm,
        drop shadow,
        left=6pt, right=6pt, top=6pt, bottom=6pt
    ]

    \textbf{=== EMPIRICAL GUIDELINES ===}

    The following guidelines are derived from empirical studies on how different types of sensitive information interact with user intent. Use these principles to make informed decisions.

    \vspace{0.5em}
    \textbf{Part A: Context-Dependent Utility -- The Hierarchy of Intent}

    User intents exhibit a rigid stratification in their dependency on sensitive context:
    \begin{itemize}[nosep]
        \item \textbf{Task Execution \& Problem Solving:} These tasks require specific details that define the solution space. Social/Relational information and Financial information can be critical constraints, while Health information is often removable if it is irrelevant noise.
        \item \textbf{Personalized Interaction:} These tasks require emotional context for establishing rapport. Health information can be important, Social/Relational information can often be replaced, and Interests/Beliefs can explain the user's motivation.
        \item \textbf{Information Acquisition \& Analysis:} These tasks focus on objective facts; most personal details are removable, Financial information is often removable, and Professional background should be kept only when directly relevant.
    \end{itemize}

    \textbf{Part B: Strategic Divergence -- Remove vs. Replace}
    \begin{itemize}[nosep]
        \item \textbf{The Poisoned Context Effect:} For factual, analytical, and calculation-heavy tasks, fabricated replacements can be more damaging than removal. Silence is superior to false context.
        \item \textbf{The Narrative Anchoring Effect:} For structural guidance or empathy-driven tasks, replacement can preserve conversational shape and tone better than deletion.
    \end{itemize}

    \textbf{Part C: Combinatorial Dynamics -- Synergy and Antagonism}
    \begin{itemize}[nosep]
        \item \textbf{Synergistic pairs:} logically interlinked details, such as Health + Interests/Beliefs, should be handled consistently because removing only one can destroy the causal narrative.
        \item \textbf{Antagonistic pairs:} redundant details, such as Social information + Behavioral data, should be protected together because keeping one can reveal the other.
    \end{itemize}

    \textbf{Part D: Decision Procedure}
    \begin{enumerate}[nosep]
        \item Use the intent label and claim metadata already provided in the Privacy Claims input. Do not re-extract the privacy claims.
        \item For each privacy item, determine whether it is a critical constraint, contextual anchor, or irrelevant noise.
        \item Choose the strategy based on intent-information interaction.
        \item Choose between remove and replace based on whether the task favors factual integrity or structural coherence.
        \item Check for synergistic or antagonistic relationships.
    \end{enumerate}

    \end{tcolorbox}
    \caption{The shared empirical guideline block referenced by the sanitization prompts as \texttt{\{EMPIRICAL\_GUIDELINES\}}.}
    \label{tab:empirical_guidelines_prompt}
\end{table*}

\begin{table*}[!ht]
    \centering
    \small
    \renewcommand{\arraystretch}{1.2}

    \begin{tcolorbox}[
        enhanced,
        title=\textbf{Shared Guideline: Replacement Guidelines},
        colframe=black!75,
        colback=gray!5,
        colbacktitle=black!75,
        fonttitle=\bfseries\large,
        boxrule=0.5mm,
        arc=1.5mm,
        drop shadow,
        left=6pt, right=6pt, top=6pt, bottom=6pt
    ]

    \textbf{=== REPLACEMENT GUIDELINES ===}

    Use natural replacements:
    \begin{itemize}[nosep]
        \item \textbf{Name} $\rightarrow$ \texttt{"someone"}, \texttt{"a colleague"}, \texttt{"a family member"}, or a culturally compatible generic name.
        \item \textbf{Location} $\rightarrow$ \texttt{"a city"}, \texttt{"a hospital"}, \texttt{"a company"}, or \texttt{"a region"}.
        \item \textbf{Date/time} $\rightarrow$ \texttt{"recently"}, \texttt{"in the past"}, or \texttt{"around that time"}.
        \item \textbf{Amount} $\rightarrow$ \texttt{"a certain amount"}, \texttt{"a limited budget"}, or \texttt{"a large amount"}.
        \item \textbf{Age} $\rightarrow$ \texttt{"a minor"}, \texttt{"a young adult"}, \texttt{"middle-aged"}, or \texttt{"older adult"}.
        \item \textbf{Relationship} $\rightarrow$ \texttt{"someone I know"}, \texttt{"a family member"}, or \texttt{"a colleague"}.
        \item \textbf{Website/URL/repository/domain} $\rightarrow$ \texttt{"a website"}, \texttt{"a repository"}, \texttt{"a domain"}, or \texttt{"an online resource"}.
        \item \textbf{Product/app/platform} $\rightarrow$ \texttt{"a product"}, \texttt{"an app"}, \texttt{"a platform"}, or \texttt{"a service"}.
        \item \textbf{Technical environment} $\rightarrow$ \texttt{"a device"}, \texttt{"an operating system"}, \texttt{"a network setup"}, or \texttt{"a technical setup"}.
        \item \textbf{Interest/topic} $\rightarrow$ \texttt{"a topic"}, \texttt{"a hobby"}, \texttt{"a media item"}, \texttt{"a technical topic"}, or \texttt{"a subject area"}.
    \end{itemize}

    \texttt{replacement} must be \texttt{null} unless strategy is \texttt{replace}.

    \end{tcolorbox}
    \caption{The shared replacement guideline block referenced by the sanitization prompts as \texttt{\{REPLACEMENT\_GUIDELINES\}}.}
    \label{tab:replacement_guidelines_prompt}
\end{table*}

\end{document}